\documentclass{article}

\usepackage[preprint]{neurips_2025}

\usepackage[utf8]{inputenc} 
\usepackage[T1]{fontenc}    
\usepackage{hyperref}       
\usepackage{url}            
\usepackage{booktabs}       
\usepackage{amsfonts}       
\usepackage{nicefrac}       
\usepackage{microtype}      
\usepackage[table]{xcolor}
\usepackage[toc]{appendix}

\usepackage{graphicx}
\usepackage{caption}
\usepackage{subcaption}
\usepackage{booktabs}
\usepackage{array}
\usepackage{enumitem}
\usepackage{amsmath,amssymb}    
\usepackage{booktabs}           
\usepackage{array}              
\usepackage{multirow}           
\usepackage{makecell}
\usepackage{verbatim}           
\usepackage{hyperref}           
\usepackage{wrapfig}
\setlist[itemize]{leftmargin=9pt}

\usepackage{xspace}
\newcommand{\method}{\textbf{\MakeUppercase{SCALE}}\xspace}

\newcommand{\squishlist}{
\begin{list}{{{\small{$\bullet$}}}}
{\setlength{\itemsep}{3pt}      \setlength{\parsep}{1pt}
\setlength{\topsep}{1pt}       \setlength{\partopsep}{0pt}
\setlength{\leftmargin}{1em} \setlength{\labelwidth}{1em}
\setlength{\labelsep}{0.5em} } }
\newcommand{\squishend}{  \end{list}  }

\title{Better Reasoning with Less Data: Enhancing VLMs Through Unified Modality Scoring}

\author{%
  Mingjie Xu \textsuperscript{1}
  Andrew Estornell\textsuperscript{2}
  Hongzheng Yang\textsuperscript{3} \\
  \textbf{Yuzhi Zhao\textsuperscript{4}} \quad
  \textbf{Zhaowei Zhu\textsuperscript{5}}
  \textbf{Qi Xuan\textsuperscript{5,6}}
  \textbf{Jiaheng Wei\textsuperscript{1} \textsuperscript{†}}
\and
  $^{1}$The Hong Kong University of Science and Technology (Guangzhou) \quad \\
  $^{2}$ByteDance Seed\quad
  $^{3}$The Chinese University of Hong Kong \quad \\
  $^{4}$City University of Hong Kong \quad
  $^{5}$BIAI-ZJUT \quad
  $^{6}$Zhejiang University of Technology 
}

\begin{document}

\maketitle

\def\thefootnote{†}\footnotetext{Corresponding to jiahengwei@hkust-gz.edu.cn.}

\begin{abstract}

The application of visual instruction tuning and other post-training techniques has significantly enhanced the capabilities of Large Language Models (LLMs) in visual understanding, enriching Vision Language Models (VLMs) with more comprehensive visual language datasets.
However, the effectiveness of VLMs is highly dependent on large-scale, high-quality datasets that ensure precise recognition and accurate reasoning.
Two key challenges hinder progress: (1) \textit{noisy alignments between images and the corresponding text}, which leads to misinterpretation, and (2) \textit{ambiguous or misleading text}, which obscures visual content.
To address these challenges, we propose \method (\textbf{S}ingle modality data quality and \textbf{C}ross modality \textbf{AL}ignment \textbf{E}valuation), a novel quality-driven data selection pipeline for VLM instruction tuning datasets. Specifically, \method integrates a cross-modality assessment framework that first assigns each data entry to its appropriate vision–language task, generates general and task-specific captions (covering scenes, objects, style, etc.), and evaluates the alignment, clarity, task rarity, text coherence, and image clarity of each entry based on the generated captions. 
Experiment results demonstrate that: (1) \method achieves impressive results with a significantly reduced dataset; 
(2) By scaling down the original instruction tuning data to less than 10\%, \method
could surpass the performance of models trained on full datasets, particularly in mathematical reasoning tasks and discipline-specific knowledge, underscoring the efficacy of our approach in refining data quality for more effective VLM instruction tuning; We also reveal that: (3) Current unimodal quality assessment methods evaluate one modality while overlook the rest, which can underestimate samples essential for specific tasks and discard the lower-quality instances that help build model robustness; (4) Appropriately generated image captions provide an efficient way to transfer the image-text multimodal task into a unified text modality. 

\end{abstract}

\section{Introduction}

Visual instruction tuning \citep{liu2024visual, li2023blip} has empowered Vision-Language Models (VLMs) to interpret visual content and follow user instructions effectively, advancing their capabilities as task-oriented assistants \citep{chen2025sharegpt4v, liu2024visual, mckinzie2025mm1, openai2024gpt4o}.
Recent works \citep{NEURIPS2023_3ef61f7e, chen2025sharegpt4v, sun2024diversity, lu2021inter, tong2024cambrian} have focused on constructing specialized instruction datasets to improve VLMs' fine-grained visual perception and expand their task coverage. 

\begin{figure}[!t]
    \centering
    \caption{Illustration of two key challenges in VLM instruction-tuning data: noisy image–text alignments (left) and ambiguous or misleading text (right). Examples are drawn from the Cambrian-7M dataset~\citep{tong2024cambrian}, with text highlighted in red indicating incorrect or misleading model responses.
    }\vspace{-0.05in}
    \includegraphics[width=0.95\linewidth]{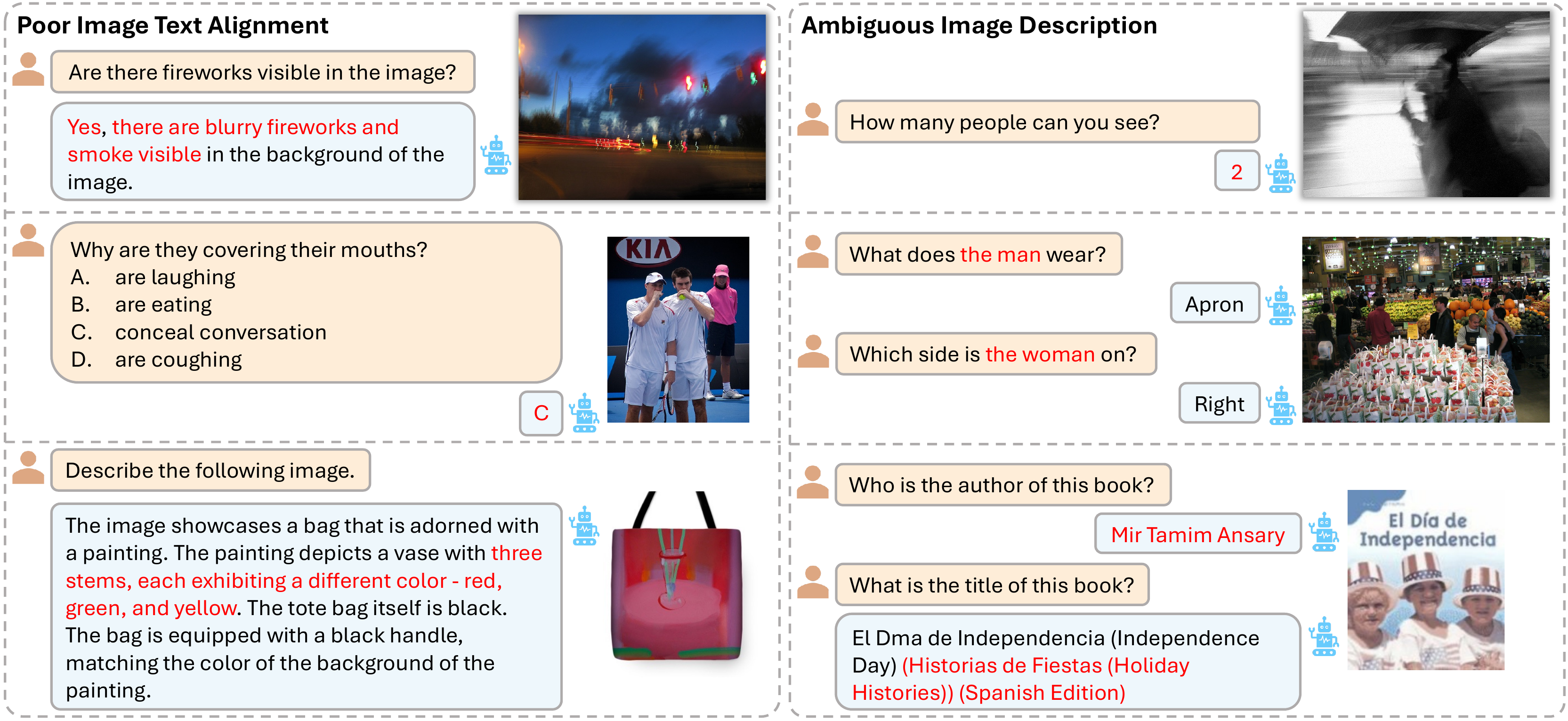}
    \vspace{-0.2in}
    \label{fig:bad-example}
\end{figure}

However, generating a sufficiently diverse and informative dataset for effective fine-tuning remains a significant challenge \citep{wei2022learning,xia2024less,liu2024automatic}. A key issue arises from the noisy image-text correspondences in existing datasets. For instance, ShareGPT4Video \citep{chen2024sharegpt4video} relies heavily on GPT-generated video captions to provide detailed descriptions, but this approach may inherit inaccuracies from GPT-4V, producing imprecise or unclear descriptions. Moreover, the cost of curating such datasets remains high. Similarly, Cambrian-1 \citep{tong2024cambrian} collected a high-quality dataset for visual instruction tuning, yet the entries still suffer from misalignment between images and their text descriptions, as well as a lack of detailed information in the image (see Figure~\ref{fig:bad-example}). This misalignment hinders the model’s understanding and generalization capabilities, especially in downstream tasks. Our approach begins with a comprehensive analysis of popular vision-language datasets, revealing two key issues (see Figure~\ref{fig:bad-example}): \textbf{(1) Noisy Image-Text Alignment}: Noisy alignments between images and corresponding text could propagate inaccurate knowledge; and \textbf{(2) Compromised Dataset Integrity}: The degradation of dataset integrity due to poor image clarity or incoherent text.
To address these challenges, we propose \method (\textbf{S}ingle modality data quality and \textbf{C}ross modality \textbf{AL}ignment \textbf{E}valuation), a novel quality-driven data selection pipeline for VLM instruction tuning datasets. \method first assesses each sample's unimodal quality using dedicated image and text rating models, then predicts its downstream task category to guide focused analysis. Next, we generate both general and task-specific captions with a lightweight VLM to unify visual and linguistic representations, enabling precise cross-model comparisons through the unified unimodality. Finally, we compute a composite score that blends image fidelity, text coherence, and multimodal alignment, and select the top-rated entries for downstream fine-tuning. 

Given the diversified data pools for instruction tuning of VLM, \method produces a curated training set that improves the quality and diversity of the dataset, leading to more robust and generalizable VLMs. We validate the effectiveness of our approach through extensive experiments and ablation studies across various benchmarks, including general VQA, image captioning, chart understanding, optical character recognition (OCR), and mathematical reasoning tasks. Notably, models fine-tuned on our selected subset not only perform competitively but also surpass full-data baselines, especially in challenging math reasoning tasks. This demonstrates the efficacy of quality-driven data curation in improving model performance. 

Our main contributions are summarized as follows:
\squishlist
\item \textbf{Automated Data Selection Pipeline:} We introduce \method, a fully automated pipeline that uses novel metrics to filter high-quality examples in instruction tuning datasets, addressing the challenges brought forward by low-quality image/text data or misalignments.
\item \textbf{Less is More:} Models trained on \method curated \textbf{10\%} subset outperform models trained on the full dataset, particularly excelling in math reasoning and discipline-specific knowledge benchmarks.
\item \textbf{New insight into Unimodal Data:} Relying solely on a unimodality for evaluation introduces biases, modality-specific metrics may overlook samples critical for robustness or mistakenly penalize valid outputs, undermining overall performance.

\item \textbf{New insight into Unified Image-Text Modality:} Appropriately generated image captions is an efficient way to transfer the image-text multi-modality alignment task into a unified text modality. 
\squishend

\section{Related Works}

\paragraph{Unimodal SFT Data Quality Evaluation}

\textbf{\textcircled{1} Image-Data Quality Evaluation.}~
In computer vision, the quality of the training data is crucial to the model performance \citep{deng2009imagenet,natarajan2013learning,liu2015classification,liu2020peer,wei2022smooth,cheng2020learning}, and several approaches have been proposed to ensure that only informative images contribute to learning. TracIn~\citep{pruthi2020estimating}, for example, quantifies the impact of individual examples by tracking loss changes during stochastic gradient descent (SGD), effectively identifying influential samples. Building on this, \citep{paul2021deep} introduce lightweight metrics, gradient norm and error L2-norm, that enable early identification of key training examples, facilitating dataset pruning without compromising performance. Similarly, MoSo~\citep{tan2024data} proposes an efficient method for pruning data by evaluating the contribution of samples to empirical risk reduction through gradient-based scoring. Expanding on these pruning techniques, RDED~\citep{sun2024diversity} improves dataset distillation by improving both realism and diversity in synthetic datasets through patch selection and stitching methods. 
\textbf{\textcircled{2} Text-Data Quality Evaluation.}~
The quality of the training text data is crucial for the performance of language models \citep{ganguli2022red,achiam2023gpt,zhuunmasking,wei2024measuring}. Previous studies have demonstrated that removing redundant or irrelevant data can significantly boost model performance. For example, the work by \citep{lee2022deduplicating, hernandez2022scaling} showed that eliminating duplicate or excessively short examples leads to notable improvements. Heuristic filtering methods, such as removing noisy or duplicate data, have similarly improved models such as T5~\citep{JMLR:v21:20-074}. 
In contrast, LESS~\citep{pmlr-v235-xia24c} introduces an influence-based approach that is aware of the optimizer that uses a low-rank gradient similarity search to select the most relevant subsets of general instruction tuning data, improving the development of specific LLM capabilities.  DS$^2$~\citep{pang2025improving} identifies systematic biases in LLM-based rating systems that degrade downstream performance. To address this, it proposes a pipeline that corrects inaccuracies in LLM-generated data quality ratings and token cleaning \citep{pang2025token} while ensuring diversity within a small, high-quality instruction-tuning dataset. In parallel, there exists a line of work in the field of information elicitation \citep{mccarthy1956measures,frongillo2015vector,gneiting2007strictly,kong2019information,liu2020incentives,wei2021sample,liu2023auditing}, quantifying the unimodal data quality without using ground-truth data. Due to space limits, we won't go through details here.

\paragraph{Multi-Modality (Image \& Text) SFT Data Quality Evaluation}

\textbf{\textcircled{1} Image-Text Matching \& Correspondence.}~
Aligning images and text is essential in multimodal learning to effectively integrate visual and textual data. CLIP~\citep{radford2021learning} achieves this by using a symmetric contrastive loss to bring matching image-text embeddings together while pushing mismatched pairs apart, resulting in unified multimodal representations. To address noisy alignments in large-scale image-text matching datasets, \citep{yang2024robust} proposes a plug-and-play regularization method that enforces proportional semantic changes across modalities, distinguishing clean from noisy pairs. Similarly, \citep{zhao2024mitigating} introduces a framework that preserves and utilizes both intra-modal and inter-modal embedding geometries to accurately filter out noisy image-text correspondences.
\textbf{\textcircled{2} Multi-Modality Data Quality Evaluation.}~
Building on the success of contrastive learning, multi-view learning has become a focal point in recent research. ALBEF~\citep{li2021align} extends this approach by generating multiple "views" of the same data using techniques such as Image-Text Contrastive (ITC) learning, Masked Language Modeling (MLM), and Momentum Distillation (MoD), enabling the model to learn robust, view-invariant representations. BLIP~\citep{pmlr-v162-li22n} further enhances this by refining multimodal alignment, while BLIP-2~\citep{li2023blip} advances the concept by transforming visual features into text-like tokens, allowing large language models to process them alongside textual inputs and improving visual-linguistic representation accuracy. Recent work by \citep{Liu_2024_CVPR} underscores the importance of clear and structured instructions for Large Vision-Language Models (LVLMs), ensuring these models fully leverage the complementary nature of visual and textual modalities, thus boosting their overall performance. Additionally, approaches like ICM-Assistant~\citep{wu2025icm} and GeoReasoner~\citep{li2024georeasoner} demonstrate the use of prior knowledge, such as predefined moderation rules and geometric data from street-view images, to maintain the quality of vision-language datasets.

\section{\texorpdfstring{The Method: \method}{The Method: SCALE}}
In this section, we will thoroughly explain our method to assess the data quality and further select the data by their retrospective quality for efficient training purposes. \method comprises two parts: multi-modality data quality metrics and a data selection pipeline to further utilize these defined metrics. in sections \ref{sec:metrics} and \ref{sec:pipeline}.

\begin{table}[h]
\vspace{-0.2in}
    \small
    \centering
    \renewcommand{\arraystretch}{1.3}
    \caption{Data‑quality metrics for unimodal (text, image) and multimodal entries. 
    }
    \begin{tabular}{p{2.2cm}p{1.2cm}p{7.8cm}} 
        \toprule
        \textbf{Metric} & \textbf{Symbol} & \textbf{Definition} \\
        \midrule
        \multicolumn{3}{l}{\textbf{Text\footnotemark[1]}} \\
        Informativeness     & \(\mathrm{INFO}\)  & Degree to which a text entry is clear, contextually relevant, and introduces novel, non‑redundant information. \\
        Complexity          & \(\mathrm{CPXT}\)  & Level of syntactic and semantic sophistication, including sentence structure variety, vocabulary richness, and conceptual depth. \\
        Completeness        & \(\mathrm{CPLT}\)  & Extent to which a text entry covers all expected or required information elements for its intended purpose. \\
        \midrule
        \multicolumn{3}{l}{\textbf{Image\footnotemark[2]}} \\
        Blur                & \(\mathrm{BLUR}\)  & Degree of loss in sharpness due to misfocus, motion, or processing artifacts. \\
        Noise               & \(\mathrm{NOISE}\) & Amount of random variation (“grain”) obscuring fine details. \\
        \midrule
        \multicolumn{3}{l}{\textbf{Multimodal\footnotemark[3]}} \\
        Clarity             & \(\mathrm{CLR}\)   & Precision and unambiguity of the text in conveying the visual content, leaving minimal room for misinterpretation. \\
        Relevance           & \(\mathrm{REL}\)   & Degree to which the textual description accurately reflects and complements the image content. \\
        Task Rarity         & \(\mathrm{TR}\)    & Novelty of the image–text pair’s underlying task, promoting coverage of uncommon or challenging scenarios. \\
        \bottomrule
    \end{tabular}\vspace{-0.1in}
    \label{tab:alt-metrics}
\end{table}

\footnotetext[1]{Text quality assessed by prompting the \href{https://huggingface.co/Qwen/Qwen2.5-32B-Instruct}{Qwen2.5-32B-Instruct} model.}
\footnotetext[2]{Image quality assessed using the \href{https://huggingface.co/zhangzicheng/q-sit}{zhangzicheng/q-sit} model.}
\footnotetext[3]{Multimodal ratings generated via \href{https://huggingface.co/Qwen/Qwen2.5-VL-7B-Instruct}{Qwen2.5-VL-7B-Instruct} (to produce captions) and then rated with \href{https://huggingface.co/Qwen/Qwen2.5-32B-Instruct}{Qwen2.5-32B-Instruct}.}

\subsection{Data‑Quality Metrics}
\label{sec:metrics}

To evaluate both unimodal 
and cross‑modal data, we introduce a set of metrics (see Table~\ref{tab:alt-metrics}), quantifying key attributes of text entries, image entries, and their combination in multimodal pairs.

\textbf{Text Quality.} We assess text entries along three dimensions: (1) \textbf{Informativeness}: Ensures that each text entry contributes substantive content rather than filler or repetition.  Without high informativeness, models may learn to generate generic or vacuous responses. (2) \textbf{Complexity}: Encourages varying syntactic and semantic structures.  By including complex sentences and rich vocabulary, we expose models to a wider linguistic range, improving their ability to handle diverse inputs. (3) \textbf{Completeness}: Guards against missing or partial information.  Incomplete entries can lead models to make unwarranted inferences or hallucinations when confronted with a piece of sparse context.

\textbf{Image Quality.} Image entries are characterized by: (1) \textbf{Blur}: Quantifies loss of detail that can impair object recognition.  By filtering out overly blurred images, we maintain a baseline of visual clarity necessary for accurate feature learning. (2) \textbf{Noise}: Measures random artifacts that may confuse both human annotators and vision models.  Controlling noise levels helps ensure that learned features correspond to real object textures rather than artifacts.

\textbf{Multimodal Quality.} For image–text pairs, we measure: (1) \textbf{Clarity}: Verifies that the text clearly describes the image.  High clarity prevents misalignment, ensuring that the model learns correct multimodal mappings rather than flawed multimodal integration. (2) \textbf{Relevance}: Confirms that text and image are semantically aligned. Otherwise, models may overfit to coincidental co‑occurrences rather than meaningful associations. (3) \textbf{Task Rarity}: Tracks how uncommon or specialized each entry’s task is.  By balancing common and rare scenarios, we avoid a skewed dataset that over-represents routine tasks, thereby improving model adaptability to novel challenges.

Together, these metrics provide a comprehensive framework for selecting and curating high‑quality samples in multimodal datasets, enhancing both model performance and generalization.

\begin{figure}[ht]
  \centering
  \vspace{-0.2in}\caption{Overview of the automated multimodal data‐selection pipeline.
  }
  
\vspace{-0.08in}\includegraphics[page=1,width=0.9\textwidth]{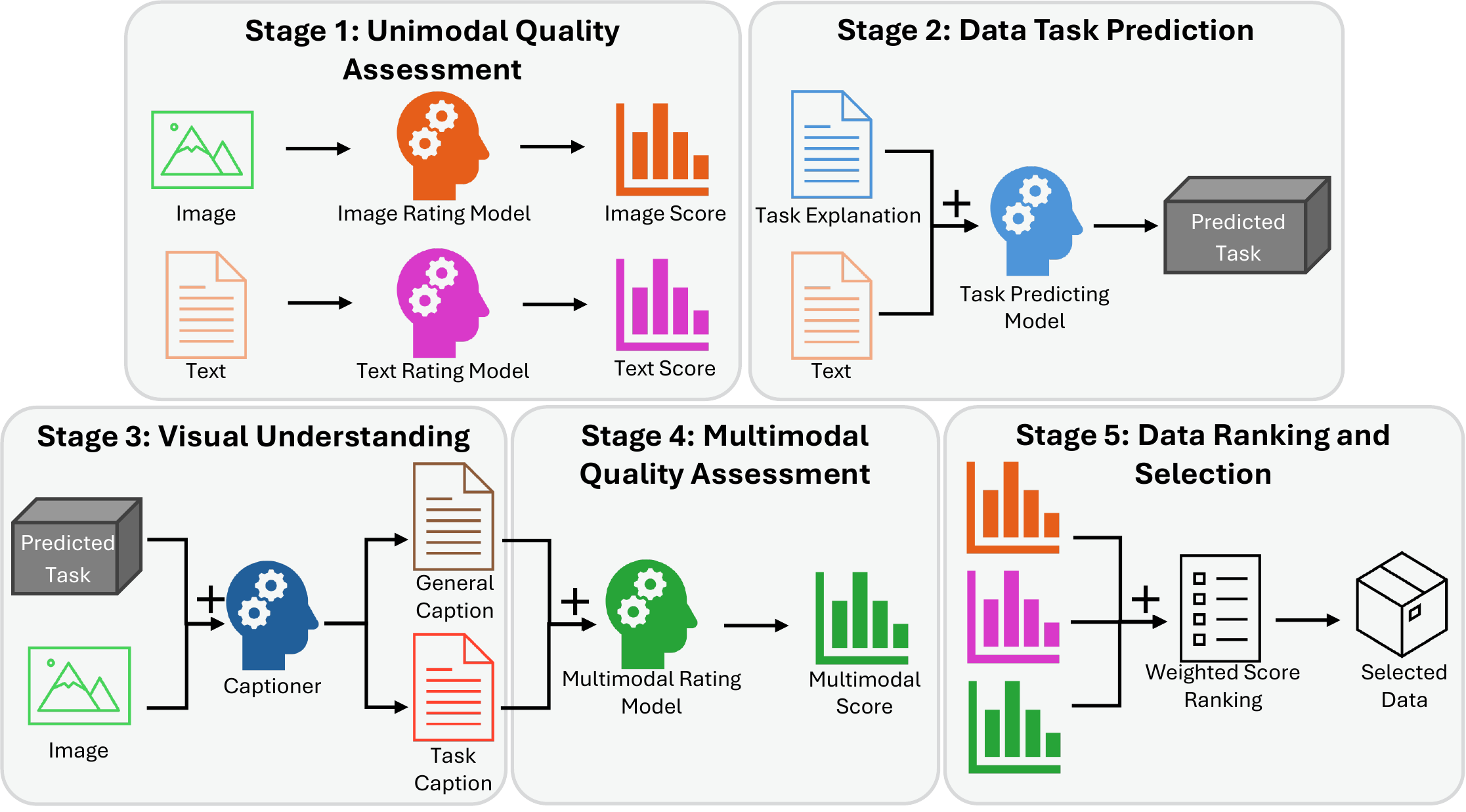}
  \label{fig:mmds-pipeline}\vspace{-0.15in}
\end{figure}
\subsection{Data Selection Pipeline}
\label{sec:pipeline}

In order to utilize our defined metrics, we develop a data selection pipeline that can rate data and select the data entry by its given score, as shown in Figure \ref{fig:mmds-pipeline}. This pipeline comprises 5 stages: the unimodal quality assessment stage, the data task prediction stage, the visual understanding stage, the multimodal quality assessment stage, and the data ranking \& selection stage.

\subsubsection{Unimodal Quality Assessment}
In the unimodal quality assessment stage, we evaluate the unimodal data quality and whether it affects the overall quality of multimodal data. With gratefully acknowledge the recent proliferation of unimodal data quality assessment techniques, which enable the application of SOTA methods for evaluating the quality of each individual modality. Our scoring module comprises two primary components: image-quality-assessment (IQA) and text-quality-assessment (TQA). 

\begin{wrapfigure}{R}{0.56\textwidth}
  \centering
  \vspace{-0.13in}\caption{Unimodal quality assessment stage illustration.}
  \includegraphics[width=\linewidth]{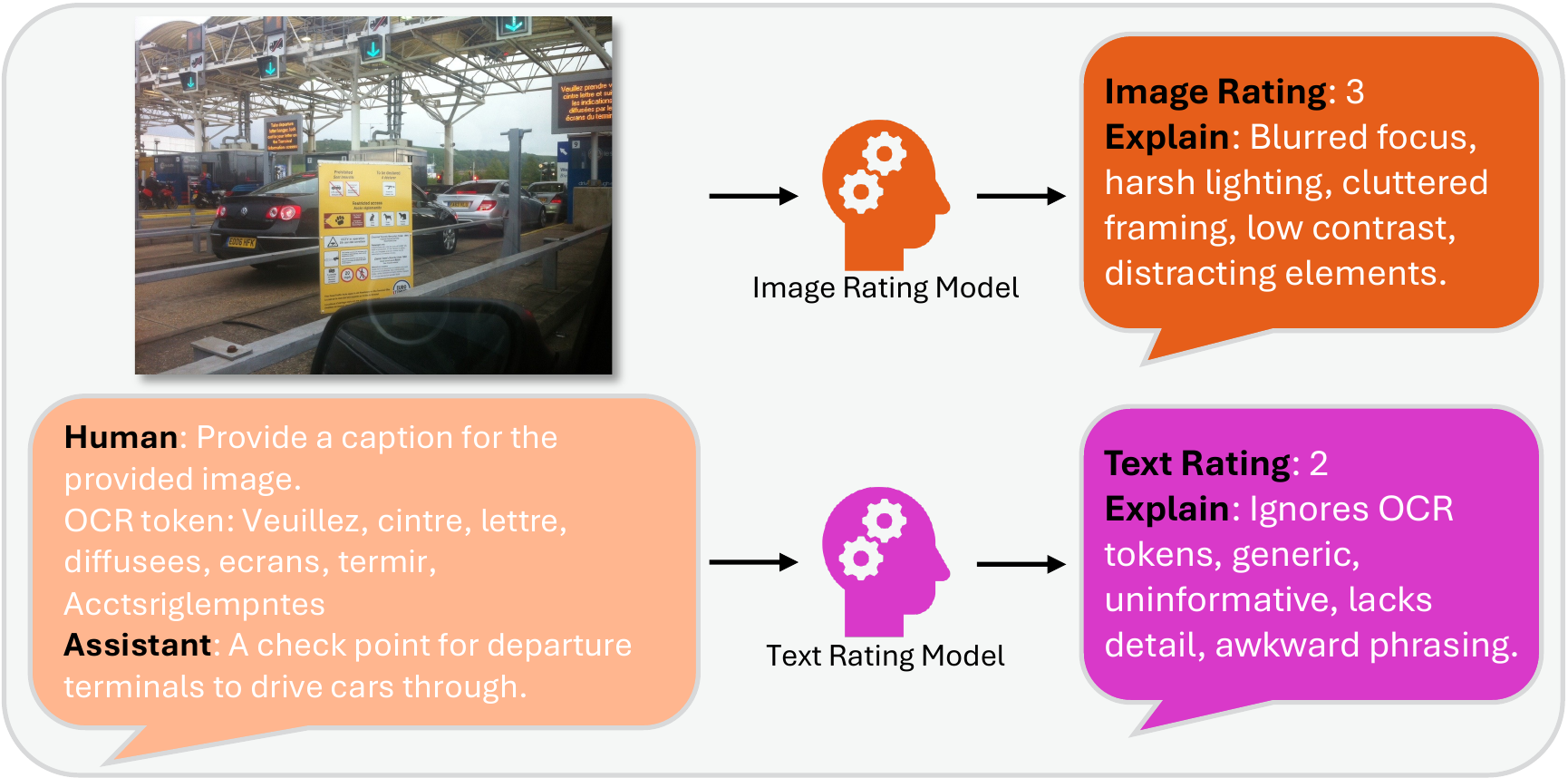}\vspace{-0.3in}
  \label{fig:stage-1-pipeline}
\end{wrapfigure}In practice, \texttt{Q-Sit}~\citep{zhang2025teaching} serves as an IQA judge to assess the image for its blurriness and noisiness. And \texttt{Qwen2.5-32-Instruct}~\citep{qwen2.5} serves as the TQA judge to test the text for its informativeness, complexity and Completeness. Both components can be formalized as follows: $S_{I} = F_{\text{IQA}}(I), S_{T} = F_{\text{TQA}}(T),$
where $I$, $T$ are the image and the text from the multimodal instruction tuning dataset, $F_{\text{IQA}}$ and $F_{\text{TQA}}$ are the IQA judge and the TQA judge.

\begin{wrapfigure}{R}{0.5\textwidth}
  \centering
  \vspace{-0.15in} \caption{Data task prediction stage illustration.}
  \includegraphics[width=\linewidth]{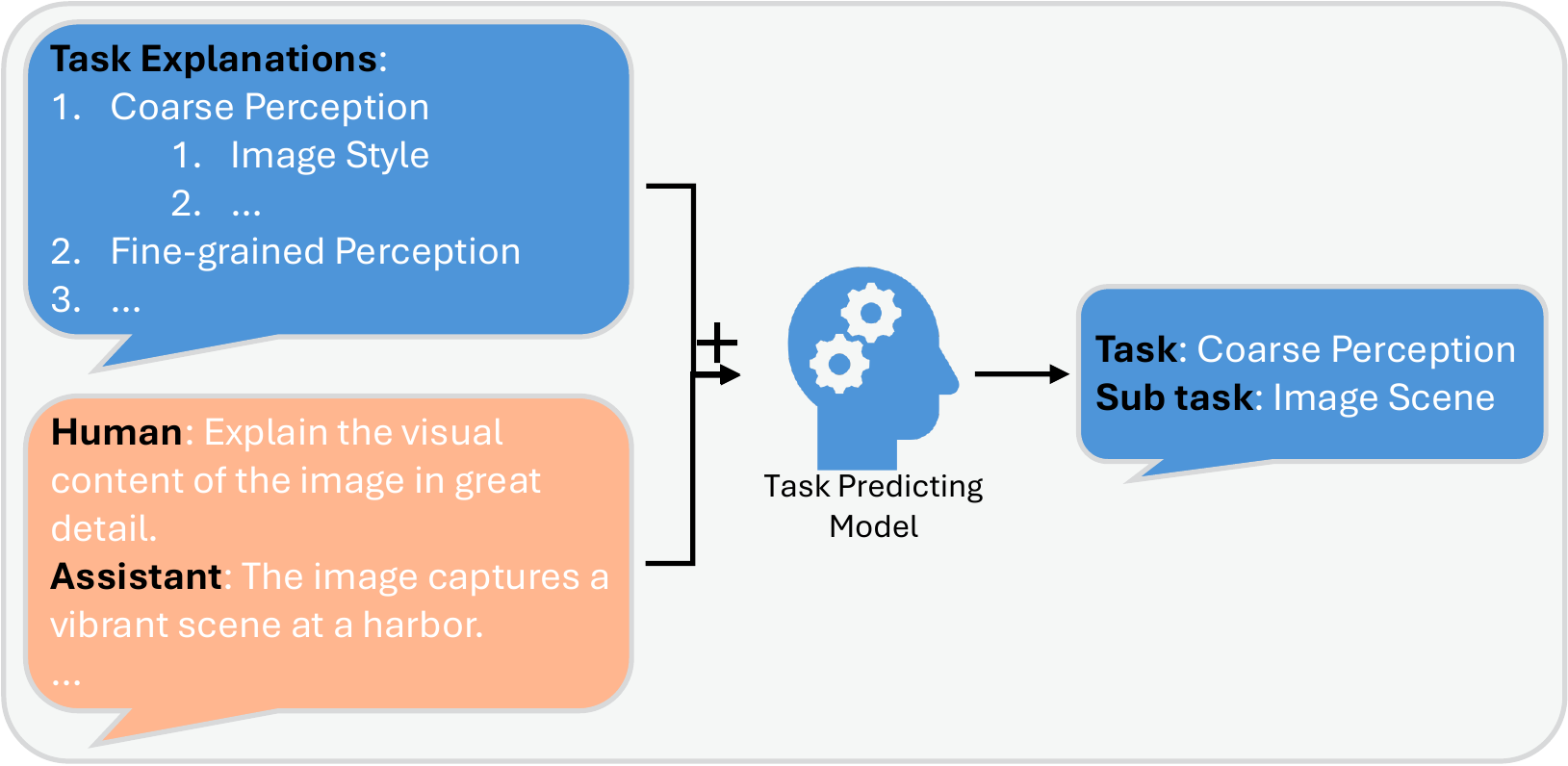}\vspace{-0.13in}
  \label{fig:stage-2-pipeline}
\end{wrapfigure}

\subsubsection{Data Task Prediction}
In order to better understand each data entry, we formalize the task it performs and apprehend its possible knowledge it contains. To achieve this, we use task categorization from MMBench~\citep{liu2025mmbench}, and we further remove tasks that are alike or included by another task. A detailed description of this task definition is provided in the supplementary material.
This data task prediction is formalized as follows: $PT=\mathbb{P}(T|\text{Task Prompt}),$
where $PT$ is the probability that the task of a data entry is specific given the text of the data entry and a list of task definitions.
Then by prompting \texttt{Qwen2.5-32B-Instruct}, it selects a task category and a sub-task category from the potential tasks pool. The prompt follows a few-shot pattern with one example per domain to calibrate the model's understanding. By parsing the model's response, we collect the task label for each data entry. Additionally, we use MMBench with its task annotations for a test run to verify the predicted tasks' accuracy against the ground truth labels. Later in Section \ref{subsec-exp}, we include Table~\ref {tab:subtask_pred}, demonstrating that this procedure can overall accurately predict the task of each data entry. This approach ensures consistent, reproducible task assignment without manual labeling, and it can adapt easily when new domains or subcategories are introduced.

\begin{figure}[h]
    \centering\vspace{-0.1in}
  \caption{Illustration of the visual understanding stage and the multimodal quality assessment stage. 
  }
  \vspace{-0.1in}
  \includegraphics[width=0.96\linewidth]{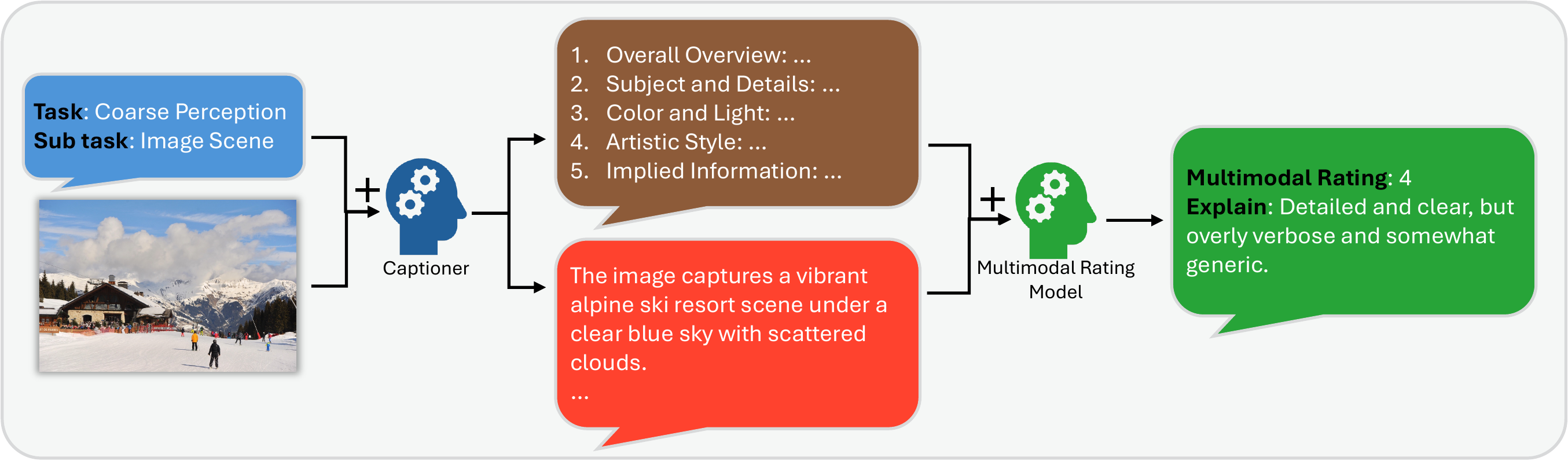}
  \vspace{-0.15in}
  \label{fig:stage-3-pipeline}
\end{figure}

\subsubsection{Visual Understanding}
In this stage, we convert the image and the text in each data entry into the same modality, in order to easily compare the difference between the image and the text, whether using Pearson Correlation or by an LLM. In order to do so, we utilize task prediction results received from the previous stage and ask a lightweight VLM~\citep{Qwen2.5-VL} targeting high-valuable aspects in the image, like targeting human interactions and objects' location in a physical relation task or object localization task. We also ask the model to generate a general caption for the image to depict its scene, environment, background atmosphere, artistic style, etc. This visual understanding process can be further formalized as: $ T_{\text{general}} = F_{\text{caption}}(I, \text{Prompt}_\text{{general}}), T_{\text{specific}} = F_{\text{caption}}(I, \text{Prompt}_{\text{specific}}).$
Here, $F_{\text{caption}}$ stands for \texttt{Qwen2.5-VL-7B} being used for generating image captions based on different needs. By giving different prompts, we ask the model to generate captions for the image, helping with the following processes.

\vspace{-0.1in}
\subsubsection{Multimodal Quality Assessment}

At this stage, we now have a unified representation of both visual and language modality, which can then be assessed by its similarity or correlation. Nevertheless, we use \texttt{Qwen2.5-32B-Instruct} as a judge to assess the quality with respect to 3 aspects. Firstly, the rating model first assesses whether the text expression is clear with respect to the image's captions, in order to avoid any ambiguous expression confound perception or reasoning knowledge. Secondly, the rating model assesses the relevance between the image's captions and the text, in order to avoid any factual errors. Thirdly, the rating model utilizes the predicted tasks with the image captions and the text to distinguish rare tasks from common ones to promote data diversity: $S_{{MM}} = F_{\text{rating}}(T_{\text{general}}, T_{\text{specific}}, T).$

\subsubsection{Data Ranking and Selection}

\begin{wrapfigure}{R}{0.66\textwidth}
  \centering
\vspace{-0.1in}  
  \caption{Ranking and selecting data stage illustration.}
  \includegraphics[width=\linewidth]{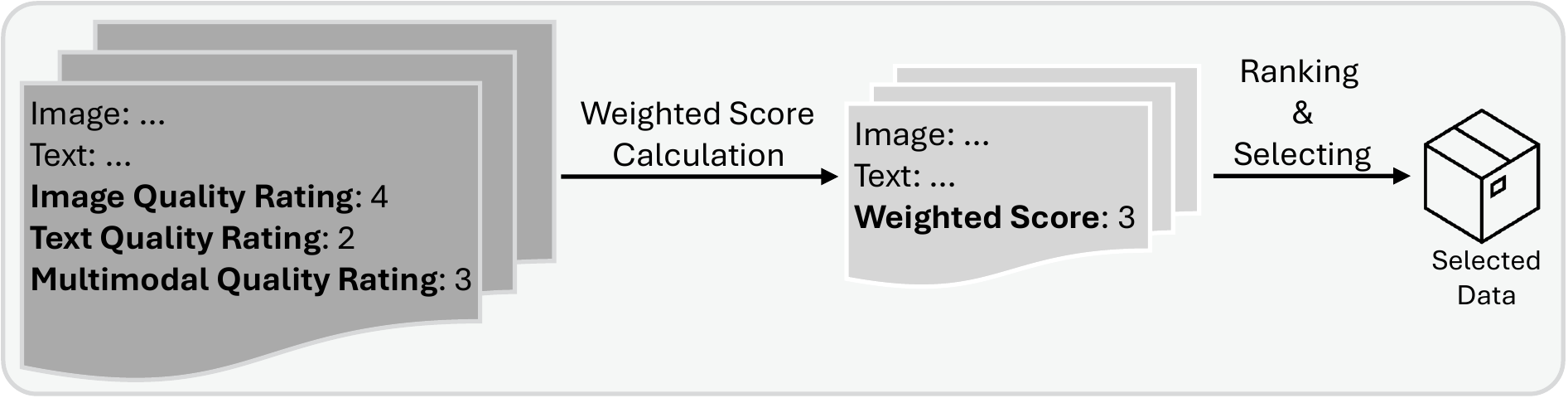}\vspace{-0.1in}
  \label{fig:stage-5-pipeline}
\end{wrapfigure}In the final stage, we compute a composite quality score for each candidate entry and retain only the highest‐ranking examples. Specifically, let \(S_I\), \(S_T\), and \(S_{MM}\) denote the image quality, text quality, and multimodal alignment scores, such that \[S_I,\,S_T,\,S_{MM} \in \{0,1,2,3,4,5\}.\] We define the overall score as $ S=0.2S_I+0.2S_T+0.6S_{MM}.$
Entries are then sorted in descending order by $S$, and the top 10 percentile are selected for downstream use, ensuring a balanced emphasis on both unimodal and multimodal quality.  

\begin{wrapfigure}{R}{0.48\textwidth}
  \centering
  \caption{Selected data's score distribution.}
  \includegraphics[width=\linewidth]{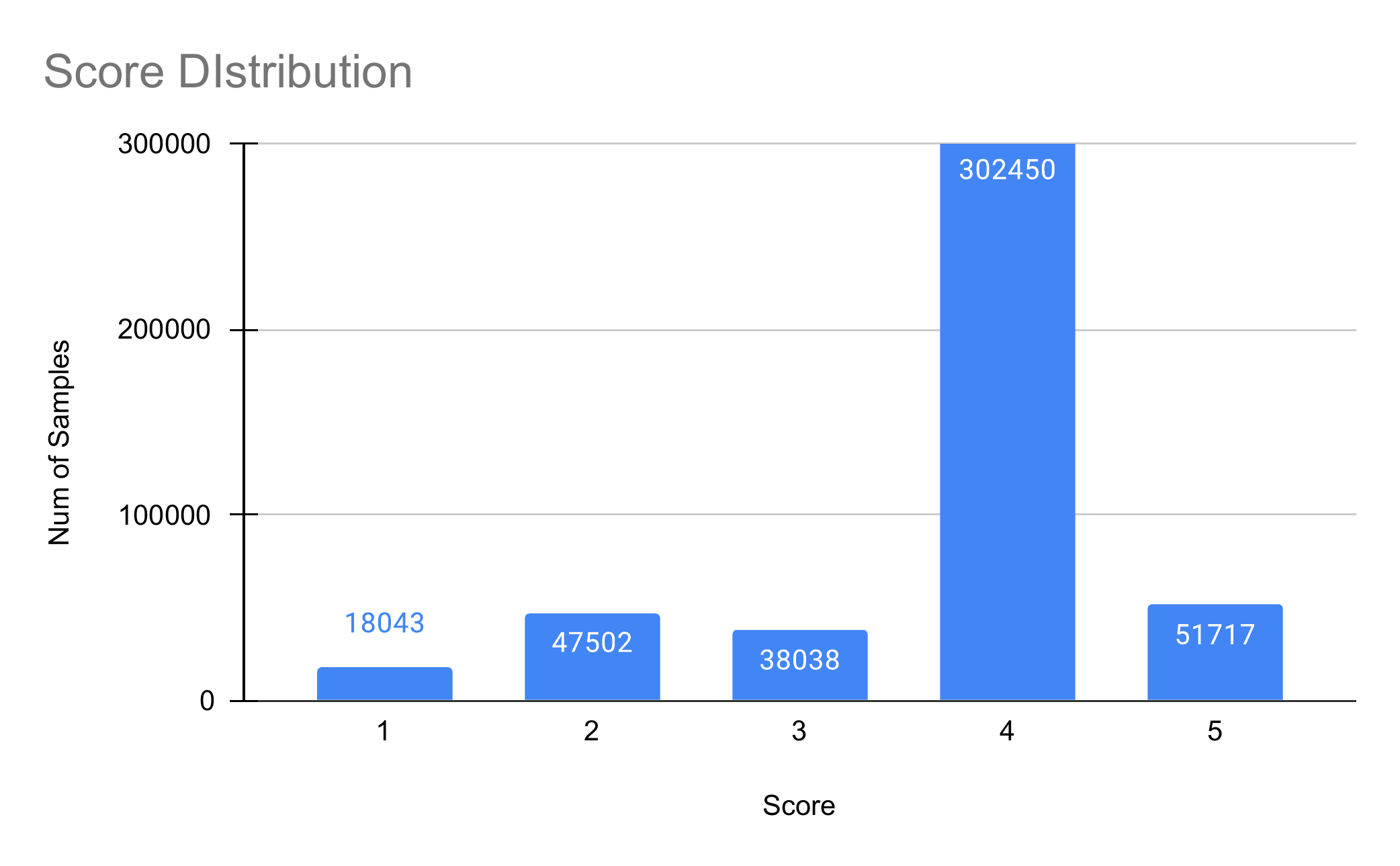}
  \label{fig:score-distri}
  \vspace{-0.4in}
\end{wrapfigure}
Figure~\ref{fig:score-distri} shows the overall quality scores produced by our data selection pipeline over the entire data pool. The vast majority of samples cluster at score 4, with roughly 302,450 entries falling into this bin. The next most frequent rating is the top score 5 (51,717 samples), whereas mid‐range scores of 2 and 3 account for about 47,502 and 38,038 entries, respectively. Only a small fraction (18,043) receive the minimum score of 1. This heavy skew toward higher ratings indicates that our ranking mechanism effectively promotes entries with strong multimodal alignment and overall fidelity, while filtering out lower‐quality candidates.

\section{Experiments}

\subsection{Experimental Setup}

\textbf{Base Models.} We use VLM Qwen2.5-VL as our base model. The model is supervised fine-tuned (SFT) on the selected data to evaluate the efficacy of different data-selection strategies.

\textbf{Baselines.} Due to the limited availability of automatic data-selection methods for multimodal datasets, we compare against two well-established baselines:  
1. \textit{Random Selection}, uniformly samples examples at random. We repeat the sampling process three times and report the mean accuracy;
2. \textit{Image quality assessment} (IQA), selects entries based on image‐quality scores produced by the \texttt{zhangzicheng/q-sit} model; higher scores correspond to better image quality;
3. \textit{Text quality assessment} (TQA), selects entries according to text‐quality ratings obtained by prompting the \texttt{Qwen/Qwen2.5-32B-Instruct} model; higher scores indicate superior text quality;
4. \textit{Joint image- and text-quality assessment} (I+T QA), selects entries based on the arithmetic mean of the image-quality score and the text-quality score;
5. \textit{CLIP Similarity}, selects examples by the cosine similarity between image and text embeddings extracted with the \texttt{openai/clip-vit-large-patch14-336} model; higher similarity denotes stronger image–text alignment.

\begin{wraptable}{R}{0.65\textwidth}  
  \centering
  \small
  \vspace{-0.16in}
    \caption{Statistics of training and validation datasets. In the tables, letter 'I' stands for image data and letter 'T' stands for text data. MC means multiple-choice.
    }
    \resizebox{0.65\textwidth}{!}{
        \begin{tabular}{l l l l l l}
        \toprule
        \textbf{Dataset} & \textbf{Size} & \textbf{Format} & \textbf{Task} & \textbf{Answer} & \textbf{Selected size} \\
        \midrule
        LLaVA-1.5-Mix \citep{Liu_2024_CVPR}       & 665K  & I+T & General QA    & Open / MC & 154K \\
        ShareGPT-4V \citep{chen2025sharegpt4v}    & 1.2M  & I+T & Caption       & Open      & 289K \\
        Geometry3K \citep{lu2021inter}            & 3K    & I+T & Mathematics   & MC        & 466      \\
        ChartQA \citep{masry-etal-2022-chartqa}   & 32K   & I+T & Chart         & Open      & 6K    \\
        InfoVQA \citep{mathew2022infographicvqa}  & 30K   & I+T & OCR           & Open      & 5K    \\
        A-OKVQA \citep{schwenk2022okvqa}          & 24K   & I+T & Knowledge     & MC        & 3K    \\
        DocVQA \citep{mathew2021docvqa}           & 50K   & I+T & Document      & Open      & 9K    \\
        AllSeeing-V2 \citep{wang2024all}          & 127K  & I+T & Grounding     & Open      & 29K   \\
        \bottomrule
      \end{tabular}
    }
    \label{tab:data-stats}
    \vspace{-0.1in}
\end{wraptable}
\textbf{Data Pool.} In Table~\ref{tab:data-stats}, we construct a data pool of 500K entries by down-sampling commonly used VLM training datasets. It spans tasks such as scene reasoning, visual recognition, and knowledge-intensive QA. Our automated data-selection pipeline is applied to this pool to produce the final training subsets.

\subsection{Experimental Results}\label{subsec-exp}

\subsubsection{Task Prediction Evaluation}

Task prediction is a fundamental component in data understanding, and it also helps value the importance of each data entry. We perform an experiment to assess the task prediction accuracy of our method using the MMBench dataset's manual label annotation, which has a total of 8664 distinct entries.

\begin{wraptable}{R}{0.5\textwidth}  
  \centering
  \renewcommand{\arraystretch}{1.2}
  \caption{The accuracy of using \texttt{Qwen2.5-32B-Instruct} to predict a data entry's task. Ground truth is from MMBench's task annotations.}
  \label{tab:subtask_pred}
  \resizebox{\linewidth}{!}{%
    \begin{tabular}{lcc}
      \toprule
      \textbf{Sub-task} & \textbf{Acc (\%)} & \textbf{Correct Predictions (total)} \\
      \midrule
      Image Style                             & 99.09 & 547 / 552 \\
      Image Scene                             & 91.52 & 842 / 920 \\
      Image Topic                             & 80.71 & 385 / 477 \\
      Object Localization                     & 82.13 & 593 / 722 \\
      Attribute Recognition                   & 81.87 & 524 / 640 \\
      Celebrity Recognition                   & 84.56 & 734 / 868 \\
      OCR                                     & 85.43 & 393 / 460 \\
      Attribute Comparison                    & 90.64 & 349 / 385 \\
      Action Recognition                      & 85.36 & 484 / 567 \\
      Physical Property                       & 80.32 & 400 / 498 \\
      Function Reasoning                      & 86.47 & 620 / 717 \\
      Social Relation                         & 98.97 & 479 / 484 \\
      Physical Relation                       & 85.19 & 317 / 357 \\
      Structured Image-Text                   & 83.63 & 552 / 660 \\
      Future Prediction                       & 88.80 & 317 / 357 \\
      \bottomrule
    \end{tabular}%
  }
\end{wraptable}
As shown in Table~\ref{tab:subtask_pred}, we observe that for all defined sub-tasks, the proposed method consistently obtains a prediction accuracy over 80\%. Notably, tasks like `Image Style', `Image Scene', `OCR', and `Social Relation' observe an average accuracy above 90\%, we presume that because these tasks have a clear cue word. On the other hand, the prediction on the rest of the tasks still witnesses a decent performance, the average accuracy reaches approximately 87\%, indicating balanced performance even for tasks with fewer samples. A closer examination of lower-scoring tasks such as `Image Topic' and `Physical Property' reveals that these tasks often require more nuanced semantic inference and contextual reasoning, where explicit textual markers are less prevalent. Overall, our Task Prediction Evaluation demonstrates the robustness of using \texttt{Qwen2.5-32B-Instruct} in classifying a wide range of vision–language tasks.

\subsubsection{Comparison with Baselines}

\definecolor{darkgreen}{RGB}{0,100,0}
\newcommand{\green}[1]{\textcolor{darkgreen}{#1}}
\definecolor{lightgray}{gray}{0.9}

\begin{table}[htb]
  \centering
  \renewcommand{\arraystretch}{1.2}
  \caption{Performance comparison with different baselines. We highlight improvements at least +0.5 points with \green{green} compared to the Full Data baseline.}
  \label{tab:model_performance}
  \resizebox{\textwidth}{!}{%
    \begin{tabular}{lccccccccc}
    \toprule
    \textbf{Model} & \textbf{A-OKVQA} & \textbf{CRPE Exist} & \textbf{CRPE Relation} & \textbf{LLaVA Wild} & \textbf{MMBench EN} & \textbf{MME} & \textbf{ScienceQA} & \textbf{SeedBench} & \textbf{Avg Score} \\
    \midrule
    Qwen2.5-VL-7B     & 86.9   & 97.1   & 76.2     & 80.2 & 82.9   & 2244.6 & 88.3      & 76.1     & 83.48 \\
    Random Selection  & 86.2   & 97.1   & 73.4     & 80.6 & 81.8   & 2245.3 & 87.6      & 74.3     & 82.65 \\
    ALBEF Retrieve    & 81.4   & 95.7   & 73.7     & 80.5 & 81.0   & 2221.5 & 84.5      & 71.0     & 80.89 \\
    BLIP Retrieve     & 81.2   & 95.7   & 73.6     & 80.0 & 80.6   & 2218.4 & 84.4      & 71.0     & 80.72 \\
    BLIP-2 Retrieve   & 81.3   & 95.7   & 73.6     & 79.7 & 81.2   & 2224.8 & 84.3      & 71.1     & 80.79 \\
    Clip Similarity   & 85.1   & 89.6   & 71.2     & 81.0 & 81.8   & 2249.6 & 89.1      & 75.1     & 81.66 \\
    Full Data (500K)  & 87.2   & \textbf{97.3}   & 76.5     & 80.6 & 83.3   & 2249.0 & 89.1      & 77.2     & 83.94 \\
    \rowcolor{lightgray}
    \method (Ours) & \textbf{87.5}   & 97.2   & \textbf{77.0}     & \textbf{81.3} & \textbf{83.4}   & \textbf{2250.1} & \textbf{89.6}      & \textbf{77.5}     & \textbf{84.23} \\
    \rowcolor{lightgray}
    $\Delta$          & +0.3   & -0.1   & \green{+0.5}     & \green{+0.7} & +0.1   & \green{+1.1} & \green{+0.5}      & +0.3     & +0.29 \\
    \bottomrule
  \end{tabular}%
  }
\end{table}

To comprehensively evaluate our proposed pipeline compared to other methods, we select evaluation datasets among multiple disciplines, including coarse-perception, fine-perception, common sense reasoning, math reasoning, code reasoning, and more. The experiments are conducted on the following datasets~\citep{schwenk2022okvqa, wang2024all, liu2024visual, liu2025mmbench, liang2024survey, lu2022learn, li2024seed} and use VLMEvalKit~\citep{duan2024vlmevalkit} as testing framework, \texttt{Qwen-2.5-32B-Instruct} as the judge model.

Table~\ref{tab:model_performance} shows that \texttt{Qwen2.5-VL-7B}, used as the base model, delivers consistently strong performance across nearly all benchmarks, achieving 86.9\% on A-OKVQA, 97.1\% on CRPE existence detection, 80.2\% on LLaVA Wild, and 88.3\% on ScienceQA, among others. First, the full data baseline uses the entire pool of 500,000 unfiltered multimodal examples, serving as an example for the impact of dataset size, only marginally outperforms the base model (e.g.\ 87.2\% vs. \ 86.9\% on A-OKVQA). In contrast, a random selection of samples yields lower results (e.g.\ 86.2\% vs.\ 86.9\% on A-OKVQA and 73.4\% vs.\ 76.2\% on CRPE Relation), underscoring the critical role of both diversity and intrinsic quality in the data pool. The CLIP-similarity baseline recovers some of this loss, improving to 81.0\% on LLaVA Wild and 2249.6 on MME compared to random sampling, but still fails to match the base model, indicating that \textbf{purely embedding-based retrieval lacks full multimodal understanding}. By contrast, our Multimodal Rating pipeline surpasses all other methods on every task,
demonstrating its superior capability to identify and select high-quality, semantically aligned examples. 
Compared to the CLIP-similarity baseline, this further proves that \textbf{well-crafted image captions are a feasible alternative to transferring vision-language tasks into a unified text task}.

\begin{table}[h]
  \centering
  \renewcommand{\arraystretch}{1.2}
  \caption{Performance comparison with different components. IQA (image-quality assessment only), TQA (text-quality assessment only), and (I+T)QA (joint image- and text-quality assessment)}
  \label{tab:ablation}
  \resizebox{\textwidth}{!}{%
    \begin{tabular}{lccccccccc}
    \toprule
    \textbf{Model} & \textbf{A-OKVQA} & \textbf{CRPE Exist} & \textbf{CRPE Relation} & \textbf{LLaVA Wild} & \textbf{MMBench EN} & \textbf{MME} & \textbf{ScienceQA} & \textbf{SeedBench} & \textbf{Avg Score} \\
    \midrule
    IQA               & 80.7   & 96.5   & 74.3     & 79.4 & 81.4   & 2235.1 & 85.2      & 71.2     & 81.07 \\
    TQA               & 81.2   & 95.3   & 72.7     & 80.1 & 81.0   & 2230.4 & 86.4      & 70.7     & 80.88 \\
    (I+T)QA           & 81.1   & 95.2   & 72.7     & 80.3 & 81.3   & 2240.1 & 86.6      & 76.2     & 81.68 \\
    \rowcolor{lightgray}
    \method (Ours) & \textbf{87.5}   & \textbf{97.2}   & \textbf{77.0}     & \textbf{81.3} & \textbf{83.4}   & \textbf{2250.1} & \textbf{89.6}      & \textbf{77.5}     & \textbf{84.23} \\
    \rowcolor{lightgray}
    \bottomrule
  \end{tabular}%
  }
\end{table}

\subsubsection{Ablation Studies}
To isolate the contribution of each modality in our selection pipeline, we decompose the full Multimodal Rating approach into three variants: IQA (image-only quality assessment), TQA (text-only quality assessment), and (I+T)QA (simple combination of image and text scores). This design allows us to examine whether relying on unimodal or naively fusing both modalities is sufficient to gauge the true quality of multimodal samples.

\begin{figure}
    \centering
    \vspace{-0.2in}
    \caption{We demonstrate some qualitative results of some incorrect ratings by using IQA or TQA score alone. This further demonstrates unimodal-only evaluation pitfalls: (1) image-only ratings can undervalue images critical for certain tasks and strip away low-quality examples needed for model robustness; (2) text-only metrics may misjudge perfectly instruction-following responses.}
    \includegraphics[width=1\linewidth]{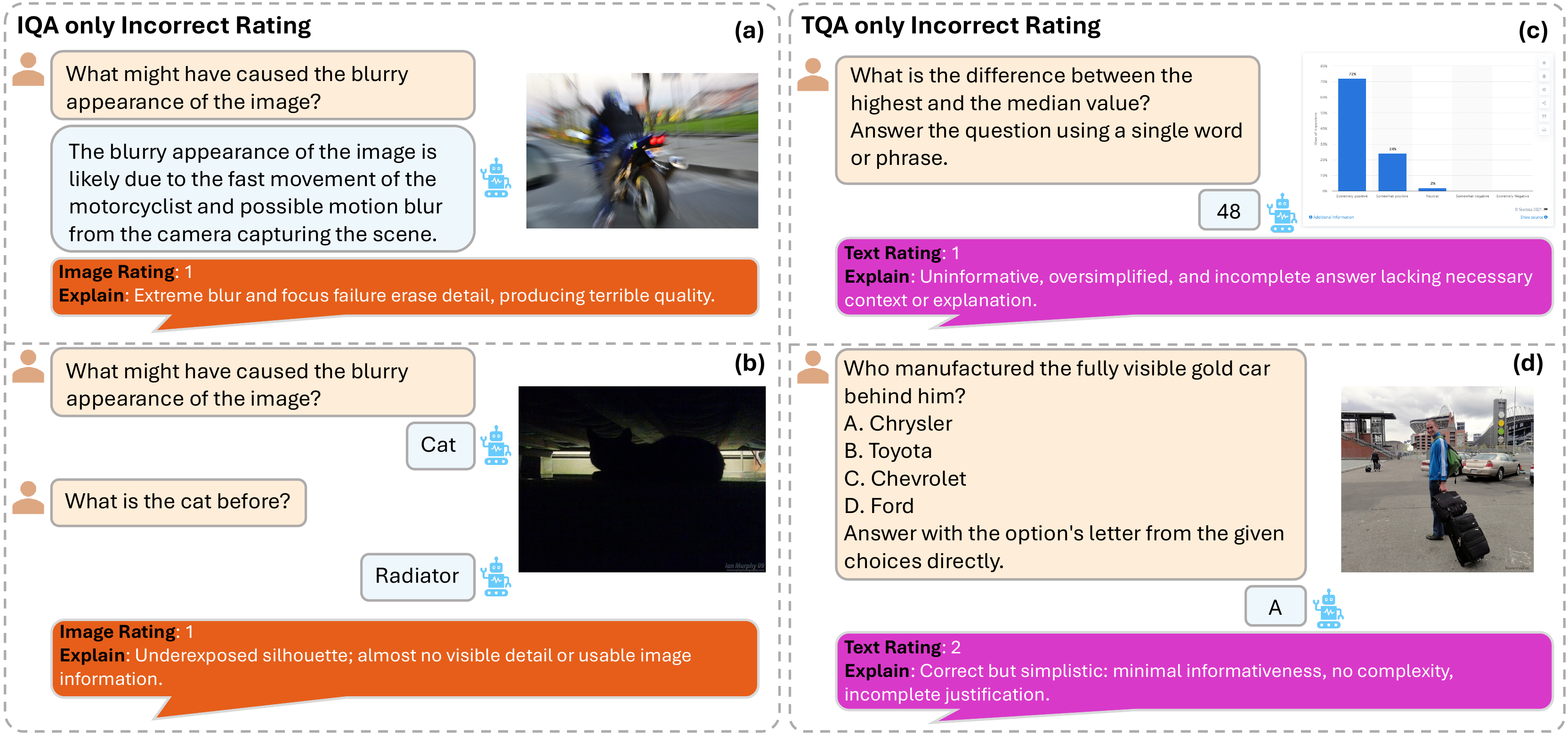}
    \label{fig:qualitative}
    \vspace{-0.3in}
\end{figure}

As reported in Table~\ref{tab:ablation}, both unimodal variants, IQA and TQA, exhibit markedly lower performance across nearly all benchmarks, e.g.\ IQA achieves only 80.7\% on A-OKVQA and 96.5\% on CRPE existence detection (versus 87.5\% and 97.2\% for the full pipeline), while TQA similarly drops to 81.2\% and 95.3\% on those tasks. The simple fusion variant (I+T)QA yields intermediate results (81.1\% on A-OKVQA, 95.2\% on CRPE Exist), yet still fails to match the full Multimodal Rating approach, demonstrating that \textbf{a straightforward summation of unimodal scores cannot capture deeper multimodal alignment}. In contrast, our complete pipeline maintains strong, consistent gains, 87.5\% on A-OKVQA, 97.2\% on CRPE Exist, 81.3\% on LLaVA Wild, 83.4\% on MMBench-EN, 2250.1 on MME, 89.6\% on ScienceQA, and 77.5\% on SeedBench, confirming that each component and its carefully designed interactions are necessary to accurately identify high-quality examples.

Overall, experiments demonstrate that unimodal evaluation methods run the risk of overlooking essential data, diminish the benefits of diversity.
As observed in Figure~\ref{fig:qualitative}:

\textcircled{1} \textbf{IQA-only evaluation neglects task-specific objectives:} in the image-quality-assessment task, a model must detect and quantify image degradations, but unimodal IQA fails to account for this diagnostic requirement (Figure~\ref{fig:qualitative} (a)); 

\textcircled{2} \textbf{Relying solely on IQA scores disregards the value of quality diversity in training:} excluding low-exposure images can hinder a model's ability to learn inference based on contour and robust visual perception (Figure~\ref{fig:qualitative} (b)); 

\textcircled{3} \textbf{TQA-only evaluation focuses exclusively on textual informativeness and coherence:} thereby overlooking the critical of instruction following accuracy in generated responses (Figure~\ref{fig:qualitative} (c) and (d)).
These qualitative findings confirm that unimodal metrics alone cannot capture the nuanced interplay between visual content and textual responses.

\section{Conclusion}

We introduced \method, a quality-driven data–selection pipeline that tackles two long-standing obstacles in vision-language instruction tuning: noisy image–text alignments and ambiguous textual descriptions, by combining single-modality quality checks with cross-modality alignment scoring. Empirically, VLMs fine-tuned on a 10 \% \method selected subset not only retain but often exceed the performance of models trained on the full data, demonstrate the ``less-is-more'' hypothesis in multi-modality data. Moreover, our study yields two practical insights. First, unimodal quality metrics, when used in isolation, can bias curation pipelines by discarding low-fidelity samples that are indispensable for robustness. Second, high-quality generated captions offer a new path to recast multimodal alignment as a unified text task, simplifying evaluation and filtering. 

\bibliographystyle{plain}
\bibliography{main}

\begin{thebibliography}{10}

\bibitem{achiam2023gpt}
Josh Achiam, Steven Adler, Sandhini Agarwal, Lama Ahmad, Ilge Akkaya, Florencia~Leoni Aleman, Diogo Almeida, Janko Altenschmidt, Sam Altman, Shyamal Anadkat, et~al.
\newblock Gpt-4 technical report.
\newblock {\em arXiv preprint arXiv:2303.08774}, 2023.

\bibitem{Qwen2.5-VL}
Shuai Bai, Keqin Chen, Xuejing Liu, Jialin Wang, Wenbin Ge, Sibo Song, Kai Dang, Peng Wang, Shijie Wang, Jun Tang, Humen Zhong, Yuanzhi Zhu, Mingkun Yang, Zhaohai Li, Jianqiang Wan, Pengfei Wang, Wei Ding, Zheren Fu, Yiheng Xu, Jiabo Ye, Xi~Zhang, Tianbao Xie, Zesen Cheng, Hang Zhang, Zhibo Yang, Haiyang Xu, and Junyang Lin.
\newblock Qwen2.5-vl technical report.
\newblock {\em arXiv preprint arXiv:2502.13923}, 2025.

\bibitem{chen2025sharegpt4v}
Lin Chen, Jinsong Li, Xiaoyi Dong, Pan Zhang, Conghui He, Jiaqi Wang, Feng Zhao, and Dahua Lin.
\newblock Sharegpt4v: Improving large multi-modal models with better captions.
\newblock In {\em European Conference on Computer Vision}, pages 370--387. Springer, 2025.

\bibitem{chen2024sharegpt4video}
Lin Chen, Xilin Wei, Jinsong Li, Xiaoyi Dong, Pan Zhang, Yuhang Zang, Zehui Chen, Haodong Duan, Zhenyu Tang, Li~Yuan, et~al.
\newblock Sharegpt4video: Improving video understanding and generation with better captions.
\newblock {\em Advances in Neural Information Processing Systems}, 37:19472--19495, 2024.

\bibitem{cheng2020learning}
Hao Cheng, Zhaowei Zhu, Xingyu Li, Yifei Gong, Xing Sun, and Yang Liu.
\newblock Learning with instance-dependent label noise: A sample sieve approach.
\newblock {\em arXiv preprint arXiv:2010.02347}, 2020.

\bibitem{deng2009imagenet}
Jia Deng, Wei Dong, Richard Socher, Li-Jia Li, Kai Li, and Li~Fei-Fei.
\newblock Imagenet: A large-scale hierarchical image database.
\newblock In {\em 2009 IEEE conference on computer vision and pattern recognition}, pages 248--255. Ieee, 2009.

\bibitem{duan2024vlmevalkit}
Haodong Duan, Junming Yang, Yuxuan Qiao, Xinyu Fang, Lin Chen, Yuan Liu, Xiaoyi Dong, Yuhang Zang, Pan Zhang, Jiaqi Wang, et~al.
\newblock Vlmevalkit: An open-source toolkit for evaluating large multi-modality models.
\newblock In {\em Proceedings of the 32nd ACM International Conference on Multimedia}, pages 11198--11201, 2024.

\bibitem{frongillo2015vector}
Rafael Frongillo and Ian~A Kash.
\newblock Vector-valued property elicitation.
\newblock In {\em Conference on Learning Theory}, pages 710--727. PMLR, 2015.

\bibitem{ganguli2022red}
Deep Ganguli, Liane Lovitt, Jackson Kernion, Amanda Askell, Yuntao Bai, Saurav Kadavath, Ben Mann, Ethan Perez, Nicholas Schiefer, Kamal Ndousse, et~al.
\newblock Red teaming language models to reduce harms: Methods, scaling behaviors, and lessons learned.
\newblock {\em arXiv preprint arXiv:2209.07858}, 2022.

\bibitem{gneiting2007strictly}
Tilmann Gneiting and Adrian~E Raftery.
\newblock Strictly proper scoring rules, prediction, and estimation.
\newblock {\em Journal of the American statistical Association}, 102(477):359--378, 2007.

\bibitem{hernandez2022scaling}
Danny Hernandez, Tom Brown, Tom Conerly, Nova DasSarma, Dawn Drain, Sheer El-Showk, Nelson Elhage, Zac Hatfield-Dodds, Tom Henighan, Tristan Hume, et~al.
\newblock Scaling laws and interpretability of learning from repeated data.
\newblock {\em arXiv preprint arXiv:2205.10487}, 2022.

\bibitem{kong2019information}
Yuqing Kong and Grant Schoenebeck.
\newblock An information theoretic framework for designing information elicitation mechanisms that reward truth-telling.
\newblock {\em ACM Transactions on Economics and Computation (TEAC)}, 7(1):1--33, 2019.

\bibitem{lee2022deduplicating}
Katherine Lee, Daphne Ippolito, Andrew Nystrom, Chiyuan Zhang, Douglas Eck, Chris Callison-Burch, and Nicholas Carlini.
\newblock Deduplicating training data makes language models better.
\newblock In {\em Proceedings of the 60th Annual Meeting of the Association for Computational Linguistics (Volume 1: Long Papers)}, pages 8424--8445, 2022.

\bibitem{li2024seed}
Bohao Li, Yuying Ge, Yixiao Ge, Guangzhi Wang, Rui Wang, Ruimao Zhang, and Ying Shan.
\newblock Seed-bench: Benchmarking multimodal large language models.
\newblock In {\em Proceedings of the IEEE/CVF Conference on Computer Vision and Pattern Recognition}, pages 13299--13308, 2024.

\bibitem{li2023blip}
Junnan Li, Dongxu Li, Silvio Savarese, and Steven Hoi.
\newblock Blip-2: Bootstrapping language-image pre-training with frozen image encoders and large language models.
\newblock In {\em International conference on machine learning}, pages 19730--19742. PMLR, 2023.

\bibitem{pmlr-v162-li22n}
Junnan Li, Dongxu Li, Caiming Xiong, and Steven Hoi.
\newblock {BLIP}: Bootstrapping language-image pre-training for unified vision-language understanding and generation.
\newblock In Kamalika Chaudhuri, Stefanie Jegelka, Le~Song, Csaba Szepesvari, Gang Niu, and Sivan Sabato, editors, {\em Proceedings of the 39th International Conference on Machine Learning}, volume 162 of {\em Proceedings of Machine Learning Research}, pages 12888--12900. PMLR, 17--23 Jul 2022.

\bibitem{li2021align}
Junnan Li, Ramprasaath Selvaraju, Akhilesh Gotmare, Shafiq Joty, Caiming Xiong, and Steven Chu~Hong Hoi.
\newblock Align before fuse: Vision and language representation learning with momentum distillation.
\newblock {\em Advances in neural information processing systems}, 34:9694--9705, 2021.

\bibitem{li2024georeasoner}
Ling Li, Yu~Ye, Bingchuan Jiang, and Wei Zeng.
\newblock Georeasoner: Geo-localization with reasoning in street views using a large vision-language model.
\newblock In {\em International Conference on Machine Learning}, pages 29222--29233. PMLR, 2024.

\bibitem{liang2024survey}
Zijing Liang, Yanjie Xu, Yifan Hong, Penghui Shang, Qi~Wang, Qiang Fu, and Ke~Liu.
\newblock A survey of multimodel large language models.
\newblock In {\em Proceedings of the 3rd International Conference on Computer, Artificial Intelligence and Control Engineering}, pages 405--409, 2024.

\bibitem{Liu_2024_CVPR}
Haotian Liu, Chunyuan Li, Yuheng Li, and Yong~Jae Lee.
\newblock Improved baselines with visual instruction tuning.
\newblock In {\em Proceedings of the IEEE/CVF Conference on Computer Vision and Pattern Recognition (CVPR)}, pages 26296--26306, June 2024.

\bibitem{liu2024visual}
Haotian Liu, Chunyuan Li, Qingyang Wu, and Yong~Jae Lee.
\newblock Visual instruction tuning.
\newblock {\em Advances in neural information processing systems}, 36, 2024.

\bibitem{liu2024automatic}
Minghao Liu, Zonglin Di, Jiaheng Wei, Zhongruo Wang, Hengxiang Zhang, Ruixuan Xiao, Haoyu Wang, Jinlong Pang, Hao Chen, Ankit Shah, et~al.
\newblock Automatic dataset construction (adc): Sample collection, data curation, and beyond.
\newblock {\em arXiv preprint arXiv:2408.11338}, 2024.

\bibitem{liu2015classification}
Tongliang Liu and Dacheng Tao.
\newblock Classification with noisy labels by importance reweighting.
\newblock {\em IEEE Transactions on pattern analysis and machine intelligence}, 38(3):447--461, 2015.

\bibitem{liu2020peer}
Yang Liu and Hongyi Guo.
\newblock Peer loss functions: Learning from noisy labels without knowing noise rates.
\newblock In {\em International conference on machine learning}, pages 6226--6236. PMLR, 2020.

\bibitem{liu2023auditing}
Yang Liu, Rixing Lou, and Jiaheng Wei.
\newblock Auditing for federated learning: A model elicitation approach.
\newblock In {\em Proceedings of the Fifth International Conference on Distributed Artificial Intelligence}, pages 1--9, 2023.

\bibitem{liu2020incentives}
Yang Liu and Jiaheng Wei.
\newblock Incentives for federated learning: A hypothesis elicitation approach.
\newblock {\em arXiv preprint arXiv:2007.10596}, 2020.

\bibitem{liu2025mmbench}
Yuan Liu, Haodong Duan, Yuanhan Zhang, Bo~Li, Songyang Zhang, Wangbo Zhao, Yike Yuan, Jiaqi Wang, Conghui He, Ziwei Liu, et~al.
\newblock Mmbench: Is your multi-modal model an all-around player?
\newblock In {\em European conference on computer vision}, pages 216--233. Springer, 2025.

\bibitem{lu2021inter}
Pan Lu, Ran Gong, Shibiao Jiang, Liang Qiu, Siyuan Huang, Xiaodan Liang, and Song-chun Zhu.
\newblock Inter-gps: Interpretable geometry problem solving with formal language and symbolic reasoning.
\newblock In {\em Proceedings of the 59th Annual Meeting of the Association for Computational Linguistics and the 11th International Joint Conference on Natural Language Processing (Volume 1: Long Papers)}, pages 6774--6786, 2021.

\bibitem{lu2022learn}
Pan Lu, Swaroop Mishra, Tanglin Xia, Liang Qiu, Kai-Wei Chang, Song-Chun Zhu, Oyvind Tafjord, Peter Clark, and Ashwin Kalyan.
\newblock Learn to explain: Multimodal reasoning via thought chains for science question answering.
\newblock {\em Advances in Neural Information Processing Systems}, 35:2507--2521, 2022.

\bibitem{masry-etal-2022-chartqa}
Ahmed Masry, Xuan~Long Do, Jia~Qing Tan, Shafiq Joty, and Enamul Hoque.
\newblock {C}hart{QA}: A benchmark for question answering about charts with visual and logical reasoning.
\newblock In Smaranda Muresan, Preslav Nakov, and Aline Villavicencio, editors, {\em Findings of the Association for Computational Linguistics: ACL 2022}, pages 2263--2279, Dublin, Ireland, May 2022. Association for Computational Linguistics.

\bibitem{mathew2022infographicvqa}
Minesh Mathew, Viraj Bagal, Rub{\`e}n Tito, Dimosthenis Karatzas, Ernest Valveny, and CV~Jawahar.
\newblock Infographicvqa.
\newblock In {\em Proceedings of the IEEE/CVF Winter Conference on Applications of Computer Vision}, pages 1697--1706, 2022.

\bibitem{mathew2021docvqa}
Minesh Mathew, Dimosthenis Karatzas, and CV~Jawahar.
\newblock Docvqa: A dataset for vqa on document images.
\newblock In {\em Proceedings of the IEEE/CVF winter conference on applications of computer vision}, pages 2200--2209, 2021.

\bibitem{mccarthy1956measures}
John McCarthy.
\newblock Measures of the value of information.
\newblock {\em Proceedings of the National Academy of Sciences}, 42(9):654--655, 1956.

\bibitem{mckinzie2025mm1}
Brandon McKinzie, Zhe Gan, Jean-Philippe Fauconnier, Sam Dodge, Bowen Zhang, Philipp Dufter, Dhruti Shah, Xianzhi Du, Futang Peng, Anton Belyi, et~al.
\newblock Mm1: methods, analysis and insights from multimodal llm pre-training.
\newblock In {\em European Conference on Computer Vision}, pages 304--323. Springer, 2025.

\bibitem{natarajan2013learning}
Nagarajan Natarajan, Inderjit~S Dhillon, Pradeep~K Ravikumar, and Ambuj Tewari.
\newblock Learning with noisy labels.
\newblock {\em Advances in neural information processing systems}, 26, 2013.

\bibitem{openai2024gpt4o}
{OpenAI}.
\newblock {GPT-4o} system card.
\newblock {\em arXiv preprint arXiv:2410.21276}, 2024.
\newblock Submitted on 25 Oct 2024.

\bibitem{pang2025token}
Jinlong Pang, Na~Di, Zhaowei Zhu, Jiaheng Wei, Hao Cheng, Chen Qian, and Yang Liu.
\newblock Token cleaning: Fine-grained data selection for llm supervised fine-tuning.
\newblock {\em arXiv preprint arXiv:2502.01968}, 2025.

\bibitem{pang2025improving}
Jinlong Pang, Jiaheng Wei, Ankit Shah, Zhaowei Zhu, Yaxuan Wang, Chen Qian, Yang Liu, Yujia Bao, and Wei Wei.
\newblock Improving data efficiency via curating {LLM}-driven rating systems.
\newblock In {\em The Thirteenth International Conference on Learning Representations}, 2025.

\bibitem{paul2021deep}
Mansheej Paul, Surya Ganguli, and Gintare~Karolina Dziugaite.
\newblock Deep learning on a data diet: Finding important examples early in training.
\newblock {\em Advances in neural information processing systems}, 34:20596--20607, 2021.

\bibitem{pruthi2020estimating}
Garima Pruthi, Frederick Liu, Satyen Kale, and Mukund Sundararajan.
\newblock Estimating training data influence by tracing gradient descent.
\newblock {\em Advances in Neural Information Processing Systems}, 33:19920--19930, 2020.

\bibitem{radford2021learning}
Alec Radford, Jong~Wook Kim, Chris Hallacy, Aditya Ramesh, Gabriel Goh, Sandhini Agarwal, Girish Sastry, Amanda Askell, Pamela Mishkin, Jack Clark, et~al.
\newblock Learning transferable visual models from natural language supervision.
\newblock In {\em International conference on machine learning}, pages 8748--8763. PMLR, 2021.

\bibitem{JMLR:v21:20-074}
Colin Raffel, Noam Shazeer, Adam Roberts, Katherine Lee, Sharan Narang, Michael Matena, Yanqi Zhou, Wei Li, and Peter~J. Liu.
\newblock Exploring the limits of transfer learning with a unified text-to-text transformer.
\newblock {\em Journal of Machine Learning Research}, 21(140):1--67, 2020.

\bibitem{schwenk2022okvqa}
Dustin Schwenk, Apoorv Khandelwal, Christopher Clark, Kenneth Marino, and Roozbeh Mottaghi.
\newblock A-okvqa: A benchmark for visual question answering using world knowledge.
\newblock In {\em European conference on computer vision}, pages 146--162. Springer, 2022.

\bibitem{sun2024diversity}
Peng Sun, Bei Shi, Daiwei Yu, and Tao Lin.
\newblock On the diversity and realism of distilled dataset: An efficient dataset distillation paradigm.
\newblock In {\em Proceedings of the IEEE/CVF Conference on Computer Vision and Pattern Recognition}, pages 9390--9399, 2024.

\bibitem{tan2024data}
Haoru Tan, Sitong Wu, Fei Du, Yukang Chen, Zhibin Wang, Fan Wang, and Xiaojuan Qi.
\newblock Data pruning via moving-one-sample-out.
\newblock {\em Advances in Neural Information Processing Systems}, 36, 2024.

\bibitem{tong2024cambrian}
Peter Tong, Ellis Brown, Penghao Wu, Sanghyun Woo, Adithya Jairam~Vedagiri IYER, Sai~Charitha Akula, Shusheng Yang, Jihan Yang, Manoj Middepogu, Ziteng Wang, et~al.
\newblock Cambrian-1: A fully open, vision-centric exploration of multimodal llms.
\newblock {\em Advances in Neural Information Processing Systems}, 37:87310--87356, 2024.

\bibitem{wang2024all}
Weiyun Wang, Yiming Ren, Haowen Luo, Tiantong Li, Chenxiang Yan, Zhe Chen, Wenhai Wang, Qingyun Li, Lewei Lu, Xizhou Zhu, et~al.
\newblock The all-seeing project v2: Towards general relation comprehension of the open world.
\newblock In {\em European Conference on Computer Vision}, pages 471--490, 2024.

\bibitem{wei2021sample}
Jiaheng Wei, Zuyue Fu, Yang Liu, Xingyu Li, Zhuoran Yang, and Zhaoran Wang.
\newblock Sample elicitation.
\newblock In {\em International Conference on Artificial Intelligence and Statistics}, pages 2692--2700. PMLR, 2021.

\bibitem{wei2022smooth}
Jiaheng Wei, Hangyu Liu, Tongliang Liu, Gang Niu, Masashi Sugiyama, and Yang Liu.
\newblock To smooth or not? when label smoothing meets noisy labels.
\newblock In {\em International Conference on Machine Learning}, pages 23589--23614. PMLR, 2022.

\bibitem{wei2024measuring}
Jiaheng Wei, Yuanshun Yao, Jean-Francois Ton, Hongyi Guo, Andrew Estornell, and Yang Liu.
\newblock Measuring and reducing llm hallucination without gold-standard answers.
\newblock {\em arXiv preprint arXiv:2402.10412}, 2024.

\bibitem{wei2022learning}
Jiaheng Wei, Zhaowei Zhu, Hao Cheng, Tongliang Liu, Gang Niu, and Yang Liu.
\newblock Learning with noisy labels revisited: A study using real-world human annotations.
\newblock In {\em International Conference on Learning Representations}, 2022.

\bibitem{wu2025icm}
Mengyang Wu, Yuzhi Zhao, Jialun Cao, Mingjie Xu, Zhongming Jiang, Xuehui Wang, Qinbin Li, Guangneng Hu, Shengchao Qin, and Chi-Wing Fu.
\newblock Icm-assistant: Instruction-tuning multimodal large language models for rule-based explainable image content moderation.
\newblock In {\em Proceedings of the AAAI Conference on Artificial Intelligence}, volume~39, pages 8413--8422, 2025.

\bibitem{xia2024less}
Mengzhou Xia, Sadhika Malladi, Suchin Gururangan, Sanjeev Arora, and Danqi Chen.
\newblock Less: Selecting influential data for targeted instruction tuning.
\newblock In {\em International Conference on Machine Learning}, pages 54104--54132. PMLR, 2024.

\bibitem{pmlr-v235-xia24c}
Mengzhou Xia, Sadhika Malladi, Suchin Gururangan, Sanjeev Arora, and Danqi Chen.
\newblock {LESS}: Selecting influential data for targeted instruction tuning.
\newblock In Ruslan Salakhutdinov, Zico Kolter, Katherine Heller, Adrian Weller, Nuria Oliver, Jonathan Scarlett, and Felix Berkenkamp, editors, {\em Proceedings of the 41st International Conference on Machine Learning}, volume 235 of {\em Proceedings of Machine Learning Research}, pages 54104--54132. PMLR, 21--27 Jul 2024.

\bibitem{qwen2.5}
An~Yang, Baosong Yang, Beichen Zhang, Binyuan Hui, Bo~Zheng, Bowen Yu, Chengyuan Li, Dayiheng Liu, Fei Huang, Haoran Wei, Huan Lin, Jian Yang, Jianhong Tu, Jianwei Zhang, Jianxin Yang, Jiaxi Yang, Jingren Zhou, Junyang Lin, Kai Dang, Keming Lu, Keqin Bao, Kexin Yang, Le~Yu, Mei Li, Mingfeng Xue, Pei Zhang, Qin Zhu, Rui Men, Runji Lin, Tianhao Li, Tingyu Xia, Xingzhang Ren, Xuancheng Ren, Yang Fan, Yang Su, Yichang Zhang, Yu~Wan, Yuqiong Liu, Zeyu Cui, Zhenru Zhang, and Zihan Qiu.
\newblock Qwen2.5 technical report.
\newblock {\em arXiv preprint arXiv:2412.15115}, 2024.

\bibitem{yang2024robust}
Yuchen Yang, Likai Wang, Erkun Yang, and Cheng Deng.
\newblock Robust noisy correspondence learning with equivariant similarity consistency.
\newblock In {\em Proceedings of the IEEE/CVF Conference on Computer Vision and Pattern Recognition}, pages 17700--17709, 2024.

\bibitem{zhang2025teaching}
Zicheng Zhang, Haoning Wu, Ziheng Jia, Weisi Lin, and Guangtao Zhai.
\newblock Teaching lmms for image quality scoring and interpreting.
\newblock {\em arXiv preprint arXiv:2503.09197}, 2025.

\bibitem{zhao2024mitigating}
Zihua Zhao, Mengxi Chen, Tianjie Dai, Jiangchao Yao, Bo~Han, Ya~Zhang, and Yanfeng Wang.
\newblock Mitigating noisy correspondence by geometrical structure consistency learning.
\newblock In {\em Proceedings of the IEEE/CVF Conference on Computer Vision and Pattern Recognition}, pages 27381--27390, 2024.

\bibitem{zhuunmasking}
Zhaowei Zhu, Jialu Wang, Hao Cheng, and Yang Liu.
\newblock Unmasking and improving data credibility: A study with datasets for training harmless language models.
\newblock In {\em The Twelfth International Conference on Learning Representations}.

\bibitem{NEURIPS2023_3ef61f7e}
Xueyan Zou, Jianwei Yang, Hao Zhang, Feng Li, Linjie Li, Jianfeng Wang, Lijuan Wang, Jianfeng Gao, and Yong~Jae Lee.
\newblock Segment everything everywhere all at once.
\newblock In A.~Oh, T.~Naumann, A.~Globerson, K.~Saenko, M.~Hardt, and S.~Levine, editors, {\em Advances in Neural Information Processing Systems}, volume~36, pages 19769--19782. Curran Associates, Inc., 2023.

\end{thebibliography}


\clearpage
\begin{appendices}
  \section*{Appendix}

\section{Limitations}

Our proposed method has demonstrated strong performance compared to other baselines, but we also acknowledge the following potential shortcomings:

\paragraph{Potential Risks of Automated Data Selection} Automated data selection methods also carry risks. Our pipeline relies on pre-trained judges (for example, \texttt{Qwen2.5-32B-Instruct}) and synthetic captions, which may reflect and amplify existing biases present in VLMs. If not carefully audited, the selection process could inadvertently underrepresent minority cultures and obscure atypical visual contexts. In safety-critical domains such as medical imaging or legal document review, misalignment between image and text or over-pruning of rare but important edge cases could lead to harmful failures.

\paragraph{Potential Misuse of the Efficient Training Paradigm} By making supervised fine-tuning of VLMs more efficient, \method may lower the barrier for malicious actors to deploy sophisticated multimodal systems for misinformation and surveillance. We therefore encourage future work to integrate bias mitigation strategies, human-in-the-loop audits, and robustness checks into data selection pipelines. By combining efficiency gains with rigorous ethical safeguards, we hope to foster a research ecosystem in which high-quality, responsible VLMs are accessible to all.

\paragraph{Computational Cost} Our approach centers on extending unimodal data‐quality assessment to the multimodal setting, with computational optimizations reserved for future work. To accommodate laboratories with limited GPU resources, the pipeline is modularized into five sequential stages, each of which can be executed independently. In future work, we will streamline these stages and refine their implementation to reduce overall GPU requirements further.


\section{Broader Impact}

\method addresses the data-quantity-quality trade-off in VLM training by showing that a 10\% curated subset can match or exceed full-scale performance. In doing so, we slash computational and carbon costs and democratize access for resource-constrained researchers.  
We highlight the following broader impacts:  

\paragraph{First Formalized Multimodal Data‐Quality Metrics and Selection Pipeline} We introduce the first end-to-end framework for quantifying and selecting data across three dimensions: textual coherence, visual fidelity, and multimodal alignment. This pipeline can be adopted in safety-critical domains such as medical imaging, legal document analysis, and autonomous navigation, to ensure that training datasets meet rigorous quality standards, thereby improving model reliability and public trust.

\paragraph{Data‐Efficient Fine-tuning on Limited Subsets} We show that models trained on only 10\% of the original data pool achieve comparable or superior performance on logic and mathematical reasoning benchmarks. By drastically reducing computing and storage requirements, our approach enables smaller institutions and edge-device deployments (e.g., mobile health diagnostics or field robotics) to benefit from large-scale multimodal models without incurring prohibitive infrastructure costs.

\paragraph{Revealing Biases in Unimodal Evaluation} We demonstrate that relying solely on image-only or text-only quality metrics can overlook critical multimodal samples and introduce systematic biases. Our multidimensional assessment framework helps prevent the exclusion of rare but important examples, such as minority dialects or atypical visual conditions, thereby fostering fairer, more inclusive models in applications like hiring, lending, and assistive technologies.

\paragraph{Image Captions as a Promising Proxy for Multimodal Alignment} We find that appropriately generated image captions effectively capture multimodal semantics, serving as a reliable proxy for image-text alignment evaluation. This insight can streamline tasks such as automated dataset labeling, alt-text generation for accessibility, and cross-modal information retrieval in digital archives and educational platforms.

\section{Additional Experiment Details}

\subsection{Experimental Setup}

In the data evaluation stage, we randomly sampled 500,000 image-text pairs from the data pool and employed a total of 60 $\times$ NVIDIA A100 (80~GB) GPUs. We first deployed the \texttt{QiT}, \texttt{Qwen-32B-Instruct}, and \texttt{Qwen-VL-7B-Instruct} models using the vLLM toolkit to support different phases of evaluation.

After completing the data evaluation, we selected approximately 50,000 image-text pairs based on the image quality score, text quality score, and overall quality score, respectively. All fine-tuning and downstream validation experiments were conducted on a server equipped with 8 $\times$ NVIDIA A100 (80~GB) GPUs.

\begin{table}[h]
  \centering
  \small
  \setlength{\tabcolsep}{4pt}
  \renewcommand{\arraystretch}{1.1}
  \resizebox{0.8\textwidth}{!}{%
    \begin{tabular}{p{3cm}p{4cm}p{6cm}}
      \toprule
      \rowcolor{white}
      \textbf{Task} & \textbf{Sub-task} & \textbf{Description} \\
      \midrule
      \multirow{3}{*}{Coarse Perception}
        & Image Style       & Classify rendering type (e.g., photograph, painting, CT scan). \\
        & Image Scene       & Identify environment (e.g., indoors, outdoors, forest, city). \\
        & Image Topic       & Determine primary subject (e.g., portrait, scenery, close-up). \\
      \midrule
      \multirow{4}{*}{\makecell[l]{Fine-grained \\ Perception \\ (single-instance)}}
        & Object Localization   & Locate and orient a single object. \\
        & Attribute Recognition & Identify object properties (shape, texture, appearance). \\
        & Celebrity Recognition & Recognize known figures or landmarks. \\
        & OCR                   & Extract text, formulas, or tables. \\
      \midrule
      \multirow{2}{*}{\makecell[l]{Fine-grained \\ Perception \\ (cross-instance)}}
        & Attribute Comparison  & Compare attributes across objects (size, color, shape). \\
        & Action Recognition    & Recognize interactions or human/object actions. \\
      \midrule
      \multirow{2}{*}{Attribute Reasoning}
        & Physical Property     & Infer material traits (e.g., fluidity, volatility). \\
        & Function Reasoning    & Predict object use or purpose (e.g., broom for sweeping). \\
      \midrule
      \multirow{2}{*}{Relation Reasoning}
        & Social Relation       & Identify human–human relationships (parent, friends). \\
        & Physical Relation     & Describe spatial relations (above, below, adjacent). \\
      \midrule
      \multirow{2}{*}{Logic Reasoning}
        & Structured Image-Text & Parse charts, diagrams, or formulas in images. \\
        & Future Prediction     & Anticipate events or state changes (weather, emotion). \\
      \bottomrule
    \end{tabular}%
  }
  \caption{Multimodal task categories, corresponding sub-tasks, and concise descriptions}
  \label{tab:domains}
\end{table}

\subsection{Multimodal Task Categories}

Based on the MMBench work, we reviewed and consolidated common multimodal data tasks. In the original MMBench categorization, some task types overlapped in purpose and scope, so we chose to keep only one representative. For example, both “Spatial Relationship” and “Physical Relation” aim to describe relationships between objects; however, “Physical Relation” more broadly encompasses ownership and interaction (e.g., one object holding or supporting another). Therefore, we retained “Physical Relation” and dropped “Spatial Relationship.”

Overall, we define the following tasks as shown in Table~\ref{tab:domains}. These tasks include entry-level visual sub-tasks, such as ``Image Style'', ``Image Scene'', and ``Image Topic'', that help the model learn and distinguish different visual characteristics and link them to downstream applications.  Other tasks build the model’s fine-grained perceptual abilities, like `Object Localization`, `Celebrity Recognition`, and `Attribute Comparison`, which in turn supply the basic information and reasoning groundwork needed for higher-level inference tasks such as `Social Relation` and `Physical Relation`.

\section{More Qualitative Results}

\begin{figure}[h]
  \centering
  \resizebox{0.9\textwidth}{!}{%
    \includegraphics{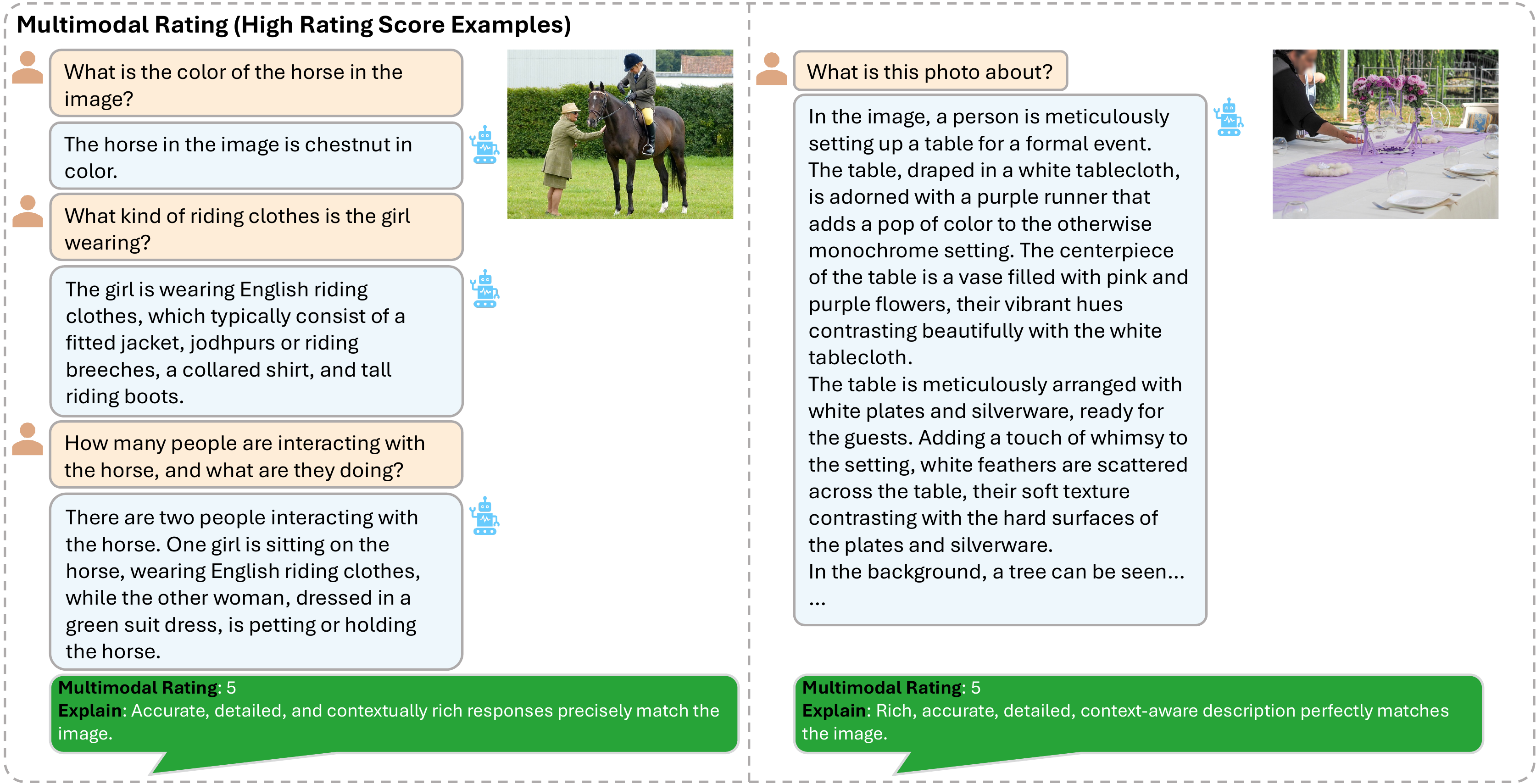}%
  }
  \caption{Examples of high‐rated multimodal data entries.}
  \label{fig:high-rating-examples}
\end{figure}

\begin{figure}[h]
    \centering
    \resizebox{0.9\textwidth}{!}{%
        \includegraphics[width=1\linewidth]{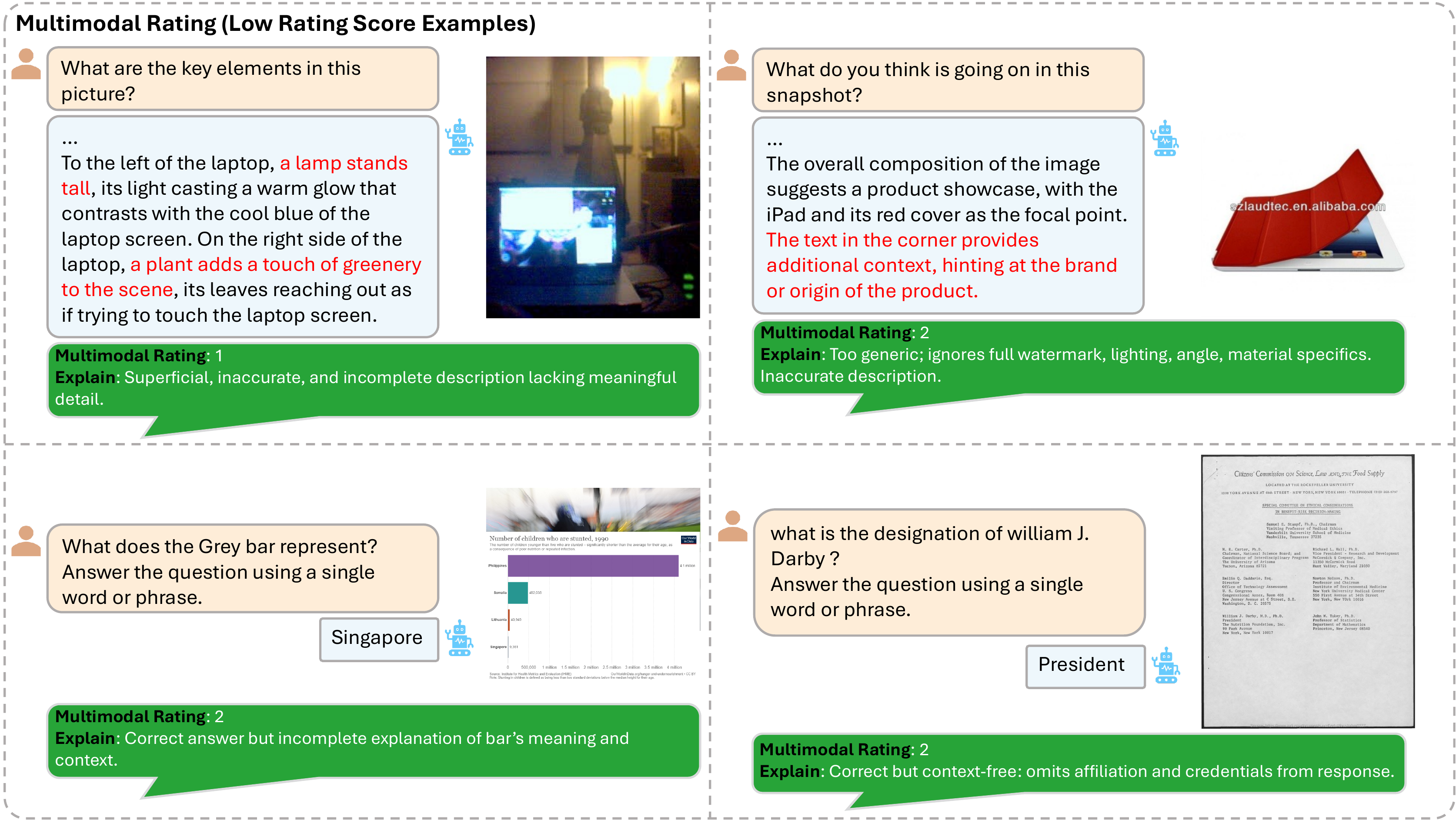}
    }
    \caption{Additional qualitative results illustrating multimodal rating score. Text highlighted in red indicates incorrect or misleading model responses. Some data entries receive low multimodal scores due to inaccuracies, while others are penalized for overly oversimplified responses.}
    \label{fig:qualitative1}
\end{figure}

\begin{figure}[h]
    \centering
    \resizebox{0.9\textwidth}{!}{%
        \includegraphics[width=1\linewidth]{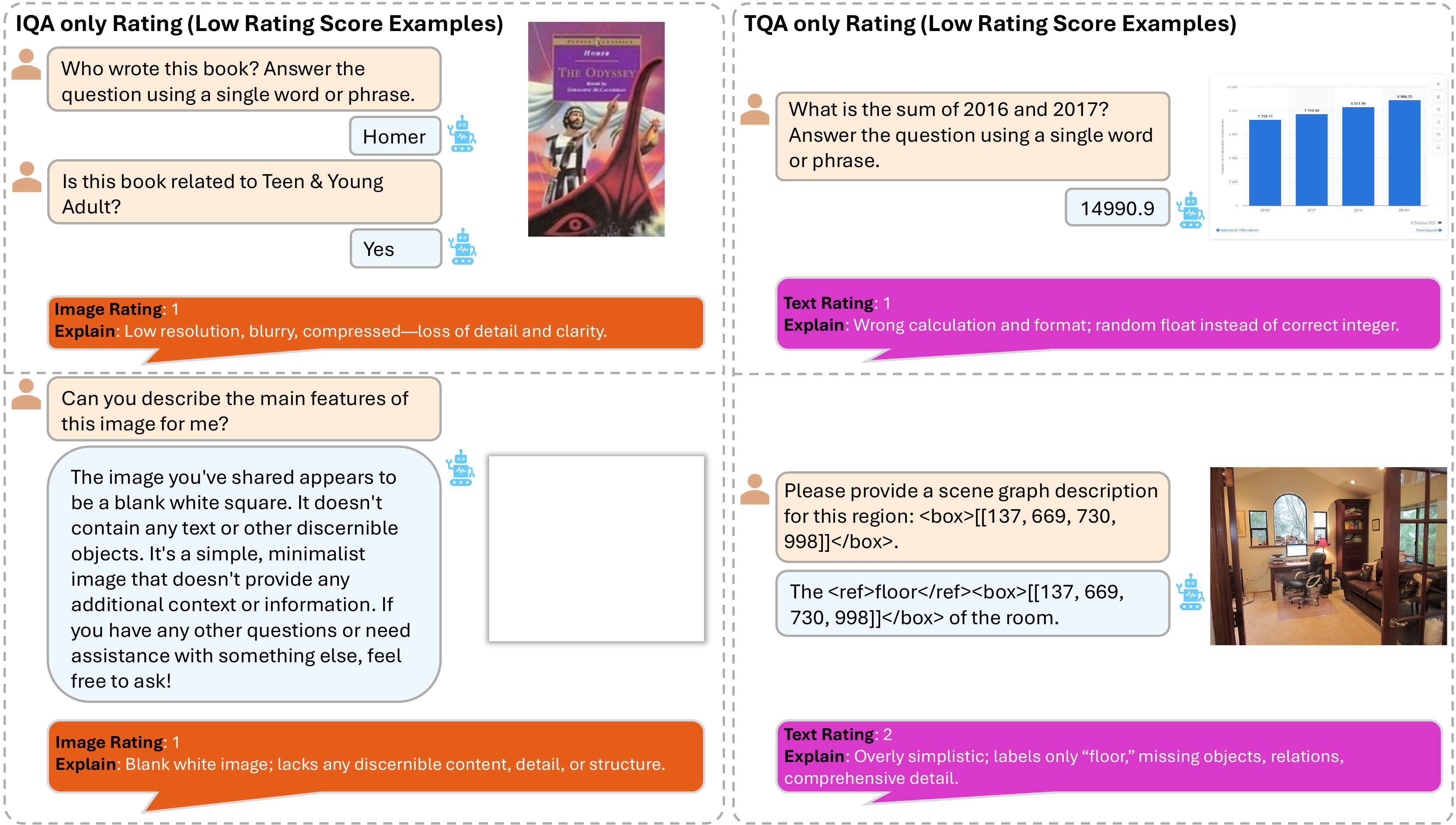}
    }
    \caption{Additional qualitative results showing image-only and text-only quality scores. These examples further demonstrate that unimodal ratings alone are insufficient to capture the true quality of a multimodal data entry.}
    \label{fig:qualitative2}
\end{figure}

\begin{figure}[h]
    \centering
    \resizebox{0.9\textwidth}{!}{%
        \includegraphics[width=1\linewidth]{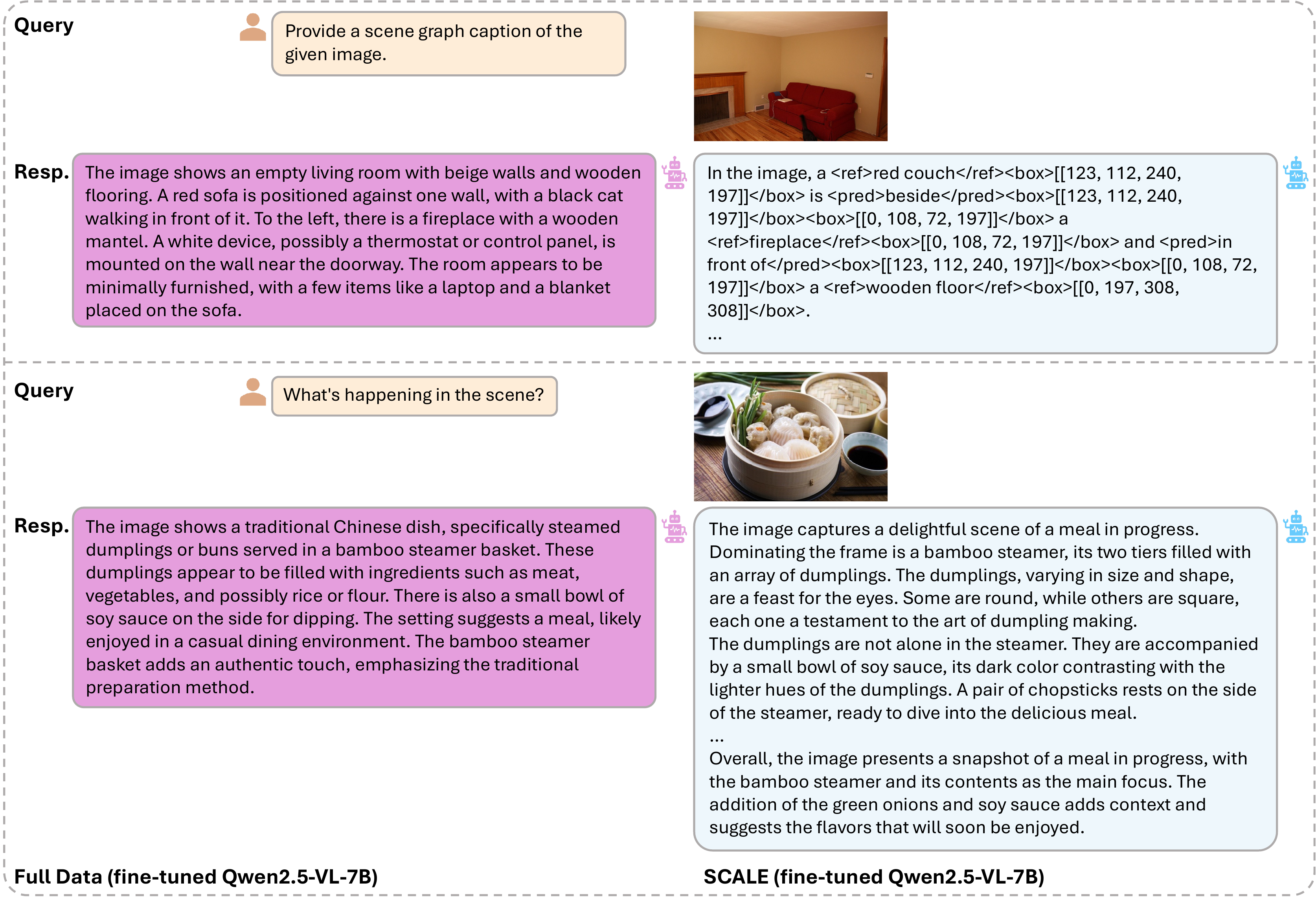}
    }
    \caption{Model responses from the full data baseline and \method. Clear differences can be observed between the two systems. In the first query, the full-data model omits the instruction and generates output without a scene graph structure, while \method correctly includes the scene graph in the required format. In the second query, \method produces a more immersive, descriptive output compared to the full-data baseline.}
    \label{fig:model-diff}
\end{figure}

We provide additional qualitative results below. The models' responses from the baseline model and \method are presented in Figure~\ref{fig:model-diff}, and we also present examples of high-quality multimodal data selected by \method in Figure~\ref{fig:high-rating-examples}. We then show the low-quality multimodal data rejected by \method in Figure~\ref{fig:qualitative1}. Finally, we illustrate examples with low IQA and TQA scores in Figure~\ref{fig:qualitative2}.

\end{appendices}

\end{document}